%% file: arxiv.tex
\documentclass{article} % For LaTeX2e
\usepackage{iclr2027_conference,times}

\input{math_commands.tex}

\usepackage[utf8]{inputenc}
\usepackage[T1]{fontenc}
\usepackage{lmodern}
\usepackage{hyperref}
\usepackage{url}
\usepackage{booktabs}
\usepackage{amsfonts}
\usepackage{amsmath}
\usepackage{nicefrac}
\usepackage{xcolor}
\usepackage{bbm}
\usepackage{enumitem}
\usepackage{amsmath}
\usepackage{amssymb}
\usepackage{amsfonts}

\usepackage{algorithm}
\usepackage{algpseudocode}

\usepackage{graphicx}
\usepackage{subcaption}

\usepackage{booktabs}
\usepackage{multirow}
\usepackage{wrapfig}

\usepackage[most]{tcolorbox}
\usepackage{xcolor}
\usepackage{listings}

\colorlet{select}{gray!20}
\definecolor{uclablue}{RGB}{0,69,126}
\definecolor{panton}{RGB}{216,27,96}
\definecolor{thupurple}{RGB}{102,0,153}
\definecolor{darkblue}{RGB}{0,0,139}
\definecolor{bigaired}{RGB}{205,92,92}

\colorlet{select}{gray!20}
\definecolor{uclablue}{RGB}{0,69,126}
\definecolor{panton}{RGB}{216,27,96}
\definecolor{thupurple}{RGB}{102,0,153}
\definecolor{darkblue}{RGB}{0,0,139}
\definecolor{bigaired}{RGB}{205,92,92}
\definecolor{sunny}{RGB}{253,184,19}  % new color for second box
\definecolor{brickred}{RGB}{203,65,84}  % frame color for second box
\newtcolorbox{llmprompt}[1][]{%
    enhanced,
    colback=select!50,
    colframe=uclablue!70,
    arc=1.5mm,
    boxrule=0.8pt,
    left=10pt, right=10pt, top=8pt, bottom=8pt,
    title={},
    listing options={%
        basicstyle=\ttfamily\small,
        breaklines=true,
        breakatwhitespace=false,
        showstringspaces=false,
        columns=flexible,
        keepspaces=true,
        upquote=true,
        frame=none,
        moredelim=[s][\color{panton}]{<}{>},
        moredelim=[s][\color{thupurple}]{<\\}{>},
        literate={User:}{{{\textcolor{darkblue}{\textbf{User:}}}}}{5}
                 {Assistant:}{{{\textcolor{bigaired}{\textbf{Assistant:}}}}}{10},
    },
    #1
}

\newtcolorbox{llmprompt2}[1][]{%
    enhanced,
    colback=sunny!30,
    colframe=brickred!70,
    arc=2mm,
    boxrule=1pt,
    left=10pt, right=10pt, top=8pt, bottom=8pt,
    title={},
    listing options={%
        basicstyle=\ttfamily\small,
        breaklines=true,
        breakatwhitespace=false,
        showstringspaces=false,
        columns=flexible,
        keepspaces=true,
        upquote=true,
        frame=none,
    },
    #1
}
\title{Beyond Binary Verifiable Rewards: Behavioral Self-Rewarding for SWE Agents}

\author{
Yiran Wu\thanks{Work partially done during an internship at Microsoft Research.} \\
Pennsylvania State University \\
\texttt{yiran.wu@psu.edu}
\And
Andrew Zhao \\
\texttt{andrewzhao112@gmail.com}
\And
Yurui Chang \\
Pennsylvania State University \\
\texttt{yuruic@psu.edu}
\And
Qingyun Wu \\
AG2AI \\
\texttt{qingyun@ag2.ai}
\And
Xuan Feng \\
Microsoft Research \\
\texttt{xuafeng@microsoft.com}
}

\iclrfinalcopy % Uncomment for camera-ready version, but NOT for submission.

\begin{document}

\maketitle

\begin{abstract}
Reinforcement learning with verifiable rewards (RLVR) provides reliable supervision for software engineering agents through executable tests, but binary rewards fail to distinguish trajectories with substantially different debugging quality. We investigate whether the current policy can judge its own trajectories and use these judgments to improve task performance without a separately trained reward model. To this end, we introduce Behavioral Self-Rewarding (BSR), a reward-refinement method that reuses the policy to compare observable debugging behavior among trajectories from the same task with the same verifier outcome. BSR aggregates pairwise preferences into task-local scores and converts them into bounded auxiliary rewards, introducing reward contrast while preserving the verifier-defined pass/fail ordering. This provides additional supervision from already-collected trajectories without requiring new environment interactions. Experiments on SWE-bench Verified and SWE-bench-Live show consistent improvements across the evaluated Qwen3-14B and Qwen3-32B configurations, supporting using the policy’s own comparative judgments as a complementary learning signal for training software engineering agents.

\end{abstract}

\begin{center}
\setlength{\tabcolsep}{1.2em}
\begin{tabular}{cccc}
    \href{https://github.com/yiranwu0/behavioral-self-rewarding}
    {\includegraphics[height=1.75em]{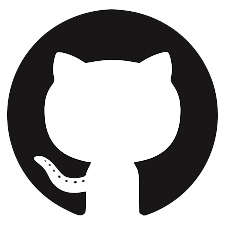}}
    &
    \href{https://yiranwu0.github.io/projects/bsr/}
    {\includegraphics[height=1.75em]{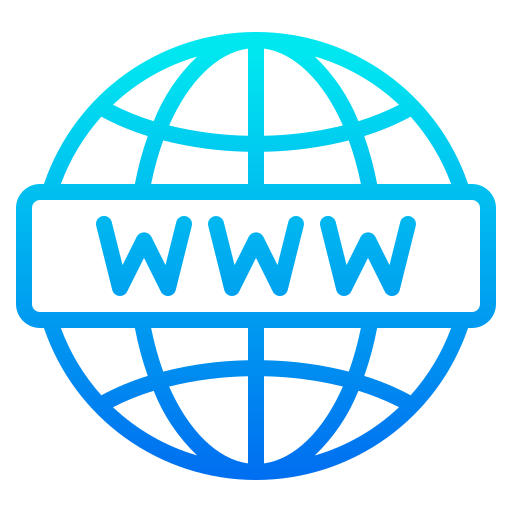}}
    &
    \href{https://wandb.ai/kevinwy00-penn-state/behavioral-self-rewarding}
    {\includegraphics[height=1.75em]{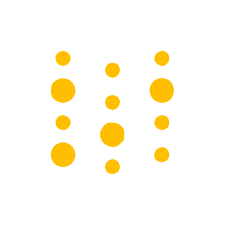}}
    &
    \href{https://huggingface.co/wyrkevin/BSR-14B}    
    {\includegraphics[height=1.75em]{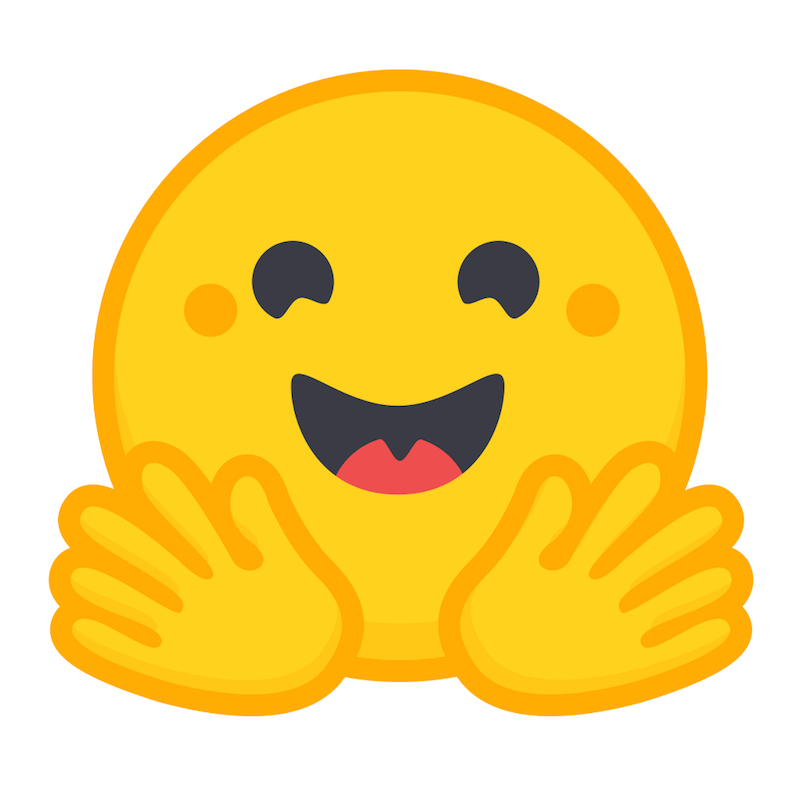}}
    \\[0.35em]

    \textbf{\scriptsize
    \href{https://github.com/yiranwu0/behavioral-self-rewarding}{Code}}
    &
    \textbf{\scriptsize
    \href{https://yiranwu0.github.io/projects/bsr/}{Project Page}}
    &
    \textbf{\scriptsize
    \href{https://wandb.ai/kevinwy00-penn-state/behavioral-self-rewarding}{Logs}}
    &
    \textbf{\scriptsize
    \href{https://huggingface.co/wyrkevin/BSR-14B}{Models}}
\end{tabular}
\end{center}

\input{method}
\input{experiment}

\vspace{-0.4em}
\section{Related Work}
\vspace{-0.3em}

\noindent\textbf{Verifiable rewards for agentic RL.}
RLVR uses automatic correctness checks, with DeepSeek-R1 demonstrating
its effectiveness at scale~\citep{guo2025deepseek}. Building on
PPO~\citep{schulman2017proximal}, GRPO removes the learned critic and
normalizes within-group rewards, GSPO uses sequence-level optimization,
and LOOP applies a leave-one-out baseline to long-horizon interactive
agents~\citep{shao2024deepseekmath,zheng2025gspo,chen2025loop}.
Hybrid-reward methods study how additional evaluative signals can
complement binary verification. HERO combines verifier outcomes with
reward-model scores through outcome-conditioned
normalization~\citep{tao2026hybrid}, while PAPO integrates rubric-based
process feedback through decoupled advantage
normalization~\citep{tan2026papo}. BSR focuses on how the additional
signal is obtained: we reuse the current policy to produce task-local
trajectory preferences, then incorporate these preferences into its
own training through bounded reward refinement.

\noindent\textbf{Software engineering agents, environments, and training.}
SWE-bench and SWE-bench-Live benchmark real-world issue
resolution~\citep{jimenez2024swebench,zhang2026swe}. SWE-agent, Agentless, and related work explore systems and workflows
for repository-level issue resolution~\citep{yang2024sweagent,
zhang2024autocoderover,xia2024agentless,wang2024openhands,
bairi2024codeplan,chen2025locagent}. SWE-Gym and
R2E-Gym provide executable repository-level training
environments~\citep{pan2025swegym,jain2025r2egym}. Training approaches
include SWE-RL on software evolution data~\citep{wei2025swerl},
multi-turn RL in stateful repositories~\citep{golubev2025training},
self-play bug injection and repair~\citep{wei2025selfplayswerl}, and
CWM's combination of interaction traces and multi-task
RL~\citep{copet2025cwm}. Methods like 
SWE-RM~\citep{shum2026swe} and SWE-TRACE~\citep{han2026swe}
develop dedicated reward models for SWE trajectories and use them
during both training and test-time inference. BSR instead investigates
whether the acting policy itself can supply useful feedback for its
own on-policy learning. Its judgments are obtained directly from the
current policy through within-task, within-outcome comparisons,
without a separate reward-model training stage. Our experiments
evaluate whether adding this self-comparison procedure improves
verifier-only training in a fixed SWE-agent setup.

\noindent\textbf{Process supervision and trajectory preferences.}
Process supervision evaluates intermediate
reasoning~\citep{lightman2023letsverify}, while trajectory preferences
support reward learning~\citep{christiano2017deep}. Bradley--Terry
aggregation and RankNet provide pairwise-ranking
foundations~\citep{bradley1952rank,burges2005learning}, and DPO learns
directly from preference pairs~\citep{rafailov2023dpo}.
LLM-as-a-judge methods provide model-generated
evaluations~\citep{zheng2023judging,liu2023geval}. Self-Rewarding Language
Models use the model's own response evaluations to construct preferences
for iterative DPO~\citep{yuan2024selfrewarding}. Process-based
Self-Rewarding applies the model's pairwise judgments to individual
mathematical reasoning steps and uses step-wise preference
optimization~\citep{zhang2025process}. BSR builds on self-generated
feedback but studies a different judgment unit and learning setting:
observable debugging behavior across complete tool-interaction
trajectories, compared within the same task and verifier-outcome
bucket. Its preferences are aggregated into bounded auxiliary rewards
for on-policy RL, rather than used as response- or step-level
preference-optimization targets.

\section{Conclusion}

We introduced Behavioral Self-Rewarding (BSR) to investigate whether
the current policy can use its own comparative judgments to improve
software-engineering performance. BSR compares trajectories from the
same task and verifier-outcome bucket, converting task-local
preferences into bounded auxiliary rewards while preserving the
verifier-defined pass/fail ordering. By reusing the policy as a
post-rollout judge, it provides additional supervision from collected
trajectories without training a separate reward model. Experiments on
SWE-bench Verified and SWE-bench-Live Lite show improvements over
verifier-only RL across the evaluated Qwen3-14B and Qwen3-32B
configurations. Together, these results support the policy's own
comparative judgments as a complement to binary verification,
allowing distinctions among otherwise reward-equivalent trajectories
to contribute to learning from repository-level interaction experience.

\bibliographystyle{plainnat}
\bibliography{ref}

\clearpage
\appendix

\input{appendix}

\end{document}

%% file: math_commands.tex
\usepackage{amsmath,amsfonts,bm}

\def\eqref#1{equation~\ref{#1}}
\def\1{\bm{1}}

\DeclareMathAlphabet{\mathsfit}{\encodingdefault}{\sfdefault}{m}{sl}
\SetMathAlphabet{\mathsfit}{bold}{\encodingdefault}{\sfdefault}{bx}{n}

%% file: method.tex
\section{Introduction}
Large language models are increasingly being deployed as software engineering
(SWE) agents: systems that operate inside real repositories and produce patches
for concrete issues through interaction with developer tools. Such
repository-level software engineering requires an agent to reason under partial
information while interacting with a changing repository and deciding how to
proceed from the observations it gathers~\citep{yang2024sweagent}. This makes
software engineering both practically important and technically demanding for
language agents, while producing rich multi-step trajectories that capture how
an agent approaches a problem. Recent benchmarks and systems have made this
setting concrete by evaluating agents on real issue-resolution tasks with
executable feedback~\citep{jimenez2024swebench,xia2024agentless,wang2024openhands}.

This setting is also a natural fit for reinforcement learning with
verifiable rewards (RLVR), where rewards come from automatic checkers
rather than learned preference models~\citep{shao2024deepseekmath,guo2025deepseek}.
Recent work extends RLVR from single-turn reasoning to multi-step
agentic RL, where models learn from trajectories of actions and
observations rather than isolated final answers~\citep{chen2025loop,
wang2025ragen,luo2025agentlightning}. Software engineering is a compelling
instance because interactions are long-horizon and behaviorally complex,
while final patches can often be validated through repository test
suites~\citep{pan2025swegym,wei2025swerl,jain2025r2egym,golubev2025training}.

However, binary test outcomes compress rich interaction trajectories into a
single success or failure signal. Failed trajectories can vary substantially
in debugging quality: some locate relevant code, gather useful evidence, or
make progress toward a fix, while others achieve little, yet both receive
reward zero. Under group-relative objectives such as GRPO or leave-one-out
PPO, rollout groups with identical verifier outcomes have zero reward variance
and therefore zero outcome-derived relative
advantages~\citep{shao2024deepseekmath,chen2025loop}. Useful distinctions among
already-collected trajectories consequently remain unavailable to the reward
signal. This is particularly costly for SWE agents, whose rollouts require
repeated model inference and tool execution. Several recent approaches enrich
verifier supervision with dense reward-model scores and learned trajectory
evaluations~\citep{tao2026hybrid,han2026swe,shum2026swe}. These approaches
demonstrate the value of supervision beyond binary outcomes, but obtain the
additional signal from a separate reward model. Rather than developing another
trajectory scorer, we investigate a complementary question:
\emph{can the current policy itself judge its own SWE trajectories, and how
can those judgments be used to improve its task performance?}

Self-rewarding and LLM-as-a-judge methods motivate this question by showing
that language models can provide evaluative signals for generated responses
and individual reasoning
steps~\citep{yuan2024selfrewarding,zheng2023judging,liu2023geval,zhang2025process}.
For interactive SWE agents, the object of judgment is instead a long-horizon
trajectory of search, inspection, editing, and verification. These trajectories
record not only the final patch but also the evidence gathered, the actions
taken in response, and the checks performed along the way. The challenge is
to turn judgments about this interaction experience into useful training
signals while retaining the distinction between passing and failing attempts.
We therefore study a restricted judging role: the executable verifier
determines the outcome of each attempt, while the current policy compares
attempts at the same task that share that outcome. This asks the policy for
local preferences, rather than an independent correctness decision or a
globally calibrated trajectory-quality score. 

To investigate this question, we introduce \emph{Behavioral Self-Rewarding}
(BSR), a reward-refinement method for multi-step agentic RLVR. For each
software-engineering task, the policy samples multiple trajectories and
receives the ordinary test-based reward. BSR partitions these trajectories
into task-local verifier-outcome buckets and reuses the current policy to
perform pairwise comparisons within each bucket. The comparisons follow a
rubric focused on observable debugging actions and their outcomes rather
than writing style or confidence. Each judgment identifies a preferred
trajectory and a coarse preference-strength label, which are aggregated
into task-local ranking scores. These scores are normalized and transformed
into centered, bounded auxiliary rewards that supplement the verifier reward.
The reward bounds ensure that every verifier-passing trajectory retains a
higher reward than every verifier-failing trajectory. BSR thus combines
current-policy trajectory comparison with verifier-conditioned reward
refinement, using the policy's preferences directly in on-policy RL without
fitting a separate trajectory reward model.

BSR integrates with group-relative RL by introducing reward contrast where
binary verification provides none. In all-fail rollout groups, nonconstant
auxiliary rewards can restore group-relative learning signal; in mixed-outcome
groups, they provide finer distinctions among trajectories sharing an outcome.
Reward refinement requires additional model inference but no new environment
interactions. We evaluate BSR against verifier-only RL to test whether this
self-judgment procedure improves the acting policy. Experiments cover
Qwen3-14B and Qwen3-32B on SWE-bench Verified, SWE-bench-Live Lite, and
Terminal-Bench 2.1. BSR improves average accuracy across the evaluated
configurations on all three benchmarks, with consistent gains in Pass@3 on
the two SWE benchmarks. Ablations examine the reward mapping and the choice
of refined outcome buckets; failure-only refinement achieves the highest
average accuracy and Pass@3 among the tested bucket configurations. In the
random-reward control, using the same auxiliary reward amplitude range
without trajectory rankings does not reproduce these gains. Together, these
results support using the current policy's comparative judgments as an
additional source of supervision for learning from repository-level
interaction experience.

\begin{figure}
    \centering
    \includegraphics[width=1\linewidth]{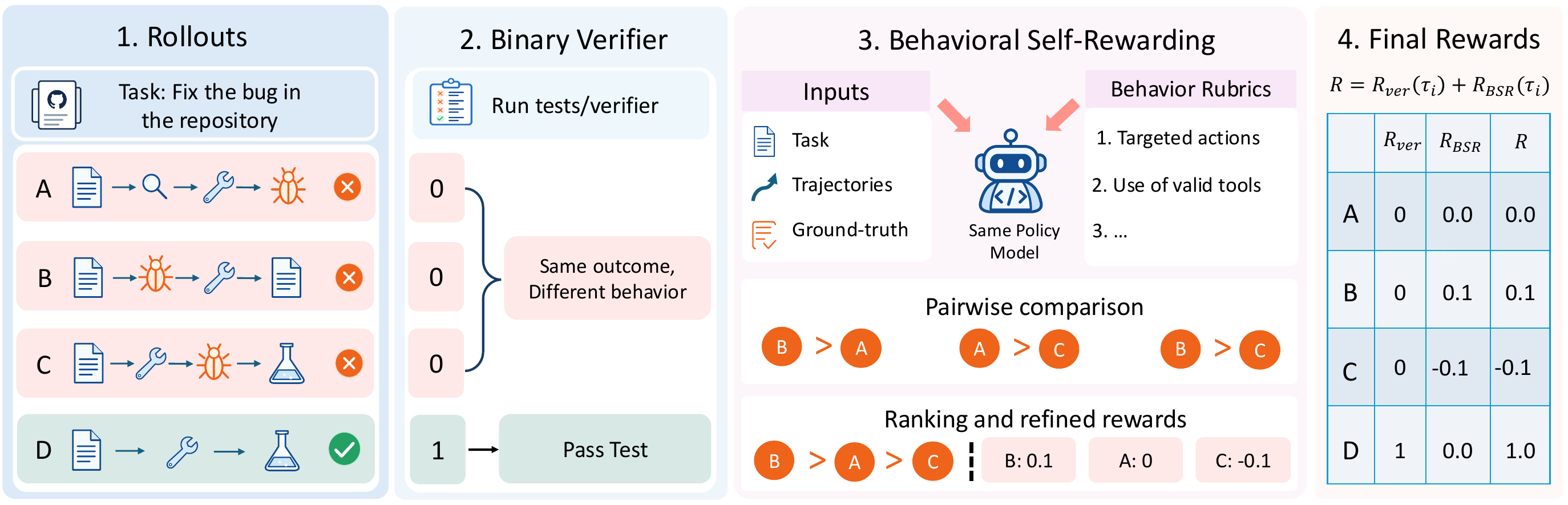}
\caption{
Overview of Behavioral Self-Rewarding (BSR). Binary rewards provide no contrast when trajectories share the same verifier outcome, while BSR ranks them within the outcome group to recover a bounded learning signal. 
}
\label{fig1}
\end{figure}

\vspace{-0.5em}
\section{Preliminaries}
\vspace{-0.5em}

We consider software-engineering tasks under RLVR: A task $x$ consists of an initial
repository state, an issue description, and an executable verifier. A policy
$\pi_\theta$ interacts with the repository through developer tools to gather
evidence, modify code, and produce a final patch. For each task, we sample
$K \geq 2$ trajectories,
\begin{equation}
\{\tau_i\}_{i=1}^{K} \sim \pi_\theta(\cdot \mid x),
\label{eq:trajectory-group}
\end{equation}
where $\tau_i$ records the full multi-step interaction with the environment.
The verifier assigns a binary reward,
\begin{equation}
R_{\mathrm{ver}}(\tau_i)
=
\begin{cases}
1, & \text{if the final patch passes the verifier,}\\
0, & \text{otherwise.}
\end{cases}
\label{eq:verifier-reward}
\end{equation}
In verifier-only RL, the training reward is $R_i = R_{\mathrm{ver}}(\tau_i)$.

Group Relative Policy Optimization (GRPO) computes advantages by normalizing
rewards within each sampled group~\citep{shao2024deepseekmath}:
\begin{equation}
A_i^{\mathrm{GRPO}}
=
\frac{R_i-\mu_R}{\sigma_R+\epsilon},
\qquad
\mu_R=\frac{1}{K}\sum_{j=1}^{K}R_j,
\qquad
\sigma_R=
\sqrt{\frac{1}{K}\sum_{j=1}^{K}(R_j-\mu_R)^2},
\label{eq:grpo-advantage}
\end{equation}
where $\epsilon>0$ ensures numerical stability. LOOP adapts
group-relative optimization to long-horizon interactive agents using a
leave-one-out baseline~\citep{chen2025loop}:

\begin{equation}
A_i^{\mathrm{LOO}}
=
R_i
-
\frac{1}{K-1}\sum_{j\ne i}R_j .
\label{eq:loo-advantage}
\end{equation}
Both methods use these advantages in clipped policy-gradient objectives without
a learned value model. With binary verifier rewards, all-fail and all-pass rollout groups have identical
rewards and therefore zero advantages under both formulations. Such groups
provide no reward-derived policy-gradient signal, even when their trajectories
differ substantially in debugging quality. This motivates refining rewards among
trajectories with the same verifier outcome while retaining verifier-defined
task success.

\section{Method}

We introduce \emph{Behavioral Self-Rewarding} (BSR) to study whether the
current policy's judgments can provide useful supervision for its own
software-engineering training. For each task, we first collect policy
trajectories and their executable verifier outcomes, then reuse the same
model as a post-rollout judge to compare attempts with the same outcome.
The judge receives the task, shared verifier outcome, compact trajectory
representations, and the training task's reference patch $p^\star$.
This deliberately asymmetric setup gives the post-rollout judge a
solution reference for interpreting the relevance of observed actions
and edits, while the acting policy generates trajectories without
access to $p^\star$ during both training and evaluation. The reference
is intended to reduce ambiguity in task-local comparisons; it is
provided as context for the behavioral rubric rather than used to
compute an explicit patch-similarity reward.
The resulting preferences are converted into auxiliary rewards for policy
optimization without training a separate reward model. The
refined training reward is
\begin{equation}
R_i = R_{\mathrm{ver}}(\tau_i) + r_i,
\label{eq:general-self-reward}
\end{equation}
where $r_i$ is a bounded auxiliary reward derived from the policy's
within-outcome comparisons. We instantiate this judgment through pairwise
ranking, asking the model to express local preferences rather than assign
absolute trajectory-quality scores. Each comparison concerns attempts at
the same task and can be inspected directly, without requiring a quality
scale calibrated across tasks. Alternative instantiations based on joint
group rankings and absolute rubric scores are discussed in
Appendix~\ref{app:alternative-behavioral-rewards}.

\subsection{Pairwise Behavioral Ranking}

BSR refines the rewards used by group-relative RL. For each task $x$, the
current policy samples $K$ trajectories, and the verifier assigns each
trajectory a binary outcome:
\begin{equation}
y_i = R_{\mathrm{ver}}(\tau_i) \in \{0,1\}.
\label{eq:verifier-outcome}
\end{equation}
We partition the trajectories into task-local verifier-outcome buckets,
\begin{equation}
B_y=\{i\in\{1,\ldots,K\}:y_i=y\},
\qquad y\in\{0,1\}.
\label{eq:outcome-buckets}
\end{equation}
Each outcome bucket has a reward amplitude $a_y\geq 0$, with $a_y=0$
disabling refinement for that outcome. The judge compares only trajectories
from the same task and verifier-outcome bucket. A failure bucket may contain
all $K$ trajectories from an all-fail rollout group or only the failed
subset of a mixed-outcome group; success buckets are handled analogously.
This restricts the policy's judging role to expressing preferences among
attempts with a shared outcome. The distinction between passing and failing
attempts is preserved by the reward-amplitude constraint in
Equation~\ref{eq:reward-ordering}.

\algrenewcommand{\algorithmiccomment}[1]{\hfill{\footnotesize$\triangleright$ #1}}

\begin{algorithm}[t]
\caption{Behavioral Self-Rewarding with Pairwise Ranking}
\label{alg:bsr-pairwise}
\small
\begin{algorithmic}[1]
\Require Task $x$, reference patch $p^\star$, policy $\pi_\theta$,
group size $K$, verifier $V$
\Require Amplitudes $a_y\ge 0$ satisfying $a_0+a_1<1$,
deadzone thresholds $\delta_y\ge 0$,
softness parameters $\tau_y>0$ for $y\in\{0,1\}$
\State Sample trajectories $\{\tau_i\}_{i=1}^{K}\sim\pi_\theta(\cdot\mid x)$
\For{$i=1,\ldots,K$}
    \State $y_i\gets V(\tau_i)$
    \Comment{$y_i=1$ if the trajectory passes the verifier; otherwise $y_i=0$}
    \State $h_i\gets\textsc{SummarizeTrajectory}(\tau_i)$
\EndFor
\For{$y\in\{0,1\}$}
    \State $B_y\gets\{i:y_i=y\}$
    \Comment{Bucket trajectories by verifier outcome}
    \If{$a_y=0$ or $|B_y|<2$}
        \State $r_i\gets0$ for all $i\in B_y$
        \State \textbf{continue}
    \EndIf
    \State $\mathcal{C}_y\gets
    \textsc{PairwiseCompare}(x,p^\star,\{h_i:i\in B_y\},B_y,y)$
    \Comment{See Algorithm~\ref{alg:pairwise-compare}}
    \State $\{r_i:i\in B_y\}\gets
    \textsc{PairwiseRewardMap}(\mathcal{C}_y,B_y,a_y,\delta_y,\tau_y)$
    \Comment{See Algorithm~\ref{alg:reward-map}}
\EndFor
\For{$i=1,\ldots,K$}
    \State $R_i\gets y_i+r_i$
    \Comment{Add bounded self-reward to verifier reward}
\EndFor
\State \Return refined rewards $\{R_i\}_{i=1}^{K}$
\end{algorithmic}
\end{algorithm}

\textbf{Outcome-separated rewarding.}
The judging criterion depends on the shared verifier outcome. Among failed
trajectories, the policy is asked to identify which attempt shows more
useful progress toward eventual success. Relevant evidence includes whether
the agent searched the appropriate part of the repository, gathered useful
evidence, made directionally plausible edits, and failed in a recoverable
way. Among successful trajectories, the policy instead compares how the
task was solved: whether the agent acted efficiently, produced a targeted
fix, avoided brittle or unnecessary changes, and verified the result.
Separate rubrics distinguish progress among unsuccessful attempts from the
quality of successful execution, rather than applying the same criterion
to both outcomes.

\textbf{Trajectory summarization.}
Raw interaction trajectories can exceed the judge's context length. We
construct compact representations by removing the thinking section from
each step, retaining exact actions, and truncating observations beyond a
token threshold. This presents the judge with the observable actions and
their outcomes. We also include deterministic trajectory statistics, such
as command executions before and after the final edit, as supporting
evidence for the comparison. The full representation and summarization
procedure are provided in
Appendix~\ref{app:trajectory-summarization}.

\textbf{Weighted pairwise comparisons.}
For each scheduled pair $(i,j)$ within a bucket, the current policy returns
a preferred trajectory $w_{ij}\in\{i,j\}$ and a coarse preference-strength
label $\alpha_{ij}\in\{1,2,3\}$, indicating slight, clear, or strong
preference. The strength label serves as a weight when fitting the
task-local scores, retaining coarse information about the relative
preference without requiring fine-grained scalar quality judgments.
To reduce positional artifacts, we randomize the order in which the two
trajectories are presented.

\subsection{From Pairwise Comparisons to Rewards}

For each outcome bucket $B_y$, let $\mathcal{C}_y$ denote the scheduled
pairwise comparisons, each represented by $(i,j,w_{ij},\alpha_{ij})$.
We aggregate these weighted preferences into task-local scores $s_i$
using a regularized weighted Bradley--Terry
model~\citep{bradley1952rank}, described in
Appendix~\ref{app:weighted-btl}. These scores summarize the policy's
comparative judgments within the bucket. To use them as auxiliary rewards,
we normalize the scores, apply a soft deadzone, and center and bound the
resulting values.

\textbf{Task-local score normalization.}
The fitted scores describe relative preferences within a task and outcome
bucket rather than a shared quality scale across tasks. For buckets with
at least two trajectories, we standardize the scores as
\begin{equation}
z_i
=
\frac{s_i-\mu_y}{\sigma_y+\epsilon},
\qquad
\mu_y=\frac{1}{|B_y|}\sum_{j\in B_y}s_j,
\qquad
\sigma_y^2=
\frac{1}{|B_y|-1}
\sum_{j\in B_y}(s_j-\mu_y)^2,
\label{eq:z-score}
\end{equation}
where $\epsilon>0$ is a small constant for numerical stability. Buckets
with fewer than two trajectories receive zero auxiliary reward because
no pairwise comparison is possible. This normalization makes the
subsequent reward map depend on relative position within the current
bucket rather than the raw scale of the fitted scores.

\textbf{Soft deadzone.}
We apply a soft deadzone to suppress normalized scores near zero and
smoothly transform larger deviations from the bucket mean:
\begin{equation}
u_i
=
\operatorname{sign}(z_i)
\tanh\!\left(
\frac{\max(|z_i|-\delta_y,0)}{\tau_y}
\right).
\label{eq:soft-deadzone}
\end{equation}
The threshold $\delta_y$ determines the region around zero in which
scores are suppressed, following the use of thresholding to remove
small deviations~\citep{drucker1997support,donoho1995denoising}.
Trajectories with $|z_i|\leq\delta_y$ have $u_i=0$. Outside this region,
the softness parameter $\tau_y$ controls how quickly $|u_i|$
approaches one.

\textbf{Centering and bounding.}
Let
\begin{equation}
\mathcal{A}_y
=
\{i\in B_y:|z_i|>\delta_y\}
\label{eq:active-set}
\end{equation}
denote the active set of trajectories outside the deadzone. If
$\mathcal{A}_y$ is empty, we set $q_i=0$ for every trajectory in the
bucket. Otherwise, we center the active scores:
\begin{equation}
\bar u_y
=
\frac{1}{|\mathcal{A}_y|}
\sum_{j\in\mathcal{A}_y}u_j,
\qquad
\tilde u_i=u_i-\bar u_y,
\quad i\in\mathcal{A}_y,
\label{eq:center-active}
\end{equation}
and rescale them by their maximum absolute value within the active set:
\begin{equation}
q_i
=
\frac{\tilde u_i}
{\max_{j\in\mathcal{A}_y}|\tilde u_j|+\epsilon},
\quad i\in\mathcal{A}_y.
\label{eq:bounded-q}
\end{equation}
For inactive trajectories $i\notin\mathcal{A}_y$, we set $q_i=0$.
The resulting scores satisfy $q_i\in[-1,1]$. Centering makes the
auxiliary signal contrastive within the active subset, while the
rescaling bounds its magnitude before applying the outcome-specific
reward amplitude.

\textbf{Final reward.}
For trajectory $i\in B_y$, the auxiliary self-judgment reward is
\begin{equation}
r_i=a_yq_i,
\label{eq:self-judge-reward}
\end{equation}
where $a_y\geq 0$ controls the amplitude for outcome bucket $y$.
Adding this refinement to the verifier reward gives
\begin{equation}
R_i
=
R_{\mathrm{ver}}(\tau_i)+r_i
=
\begin{cases}
1+a_1q_i, & R_{\mathrm{ver}}(\tau_i)=1,\\
a_0q_i, & R_{\mathrm{ver}}(\tau_i)=0.
\end{cases}
\label{eq:final-bsr-reward}
\end{equation}
Since $q_i\in[-1,1]$, passing trajectories receive rewards of at least
$1-a_1$, whereas failing trajectories receive rewards of at most $a_0$.
We choose nonnegative amplitudes satisfying
\begin{equation}
a_0+a_1<1,
\label{eq:reward-ordering}
\end{equation}
which guarantees that every verifier-passing trajectory receives a
higher reward than every verifier-failing trajectory. The refined rewards
are then used in group-relative policy optimization. In all-fail rollout
groups, nonconstant auxiliary rewards can restore group-relative learning
signal; in mixed-outcome groups, they provide additional distinctions
within each refined outcome bucket.

\begin{algorithm}[t]
\caption{\textsc{PairwiseRewardMap}}
\label{alg:reward-map}
\small
\begin{algorithmic}[1]
\Require Comparisons $\mathcal{C}_y$, bucket $B_y$, amplitude $a_y$,
deadzone $\delta_y$, softness $\tau_y$

\State $\{s_i:i\in B_y\}\gets
\textsc{WeightedBTLFit}(\mathcal{C}_y,B_y)$
\Comment{Aggregate pairwise judgments}

\State Compute $\{z_i:i\in B_y\}$ using Equation~\ref{eq:z-score}
\Comment{Normalize within the bucket}

\For{$i\in B_y$}
    \State $u_i\gets
    \operatorname{sign}(z_i)
    \tanh\!\left(
    \max(|z_i|-\delta_y,0)/\tau_y
    \right)$
    \Comment{Apply deadzone and soft mapping}
\EndFor

\State $\mathcal{A}_y\gets\{i\in B_y:|z_i|>\delta_y\}$
\Comment{Select trajectories outside the deadzone}

\If{$\mathcal{A}_y=\emptyset$}
    \State \Return $r_i=0$ for all $i\in B_y$
    \Comment{No scores outside the deadzone}
\EndIf

\State Center active scores using Equation~\ref{eq:center-active}
\Comment{Make the signal contrastive}

\State Normalize active scores using Equation~\ref{eq:bounded-q}
\Comment{Bound $q_i$ to $[-1,1]$}

\State \Return $r_i=a_yq_i$ for all $i\in B_y$
\Comment{Inactive trajectories have $q_i=0$}

\end{algorithmic}
\end{algorithm}

%% file: experiment.tex
% \vspace{-1.3em}
\section{Experiments}
% \vspace{-0.5em}

\paragraph{Setup.}
We use RLLM~\citep{rllm2025} to train on a filtered version of
R2E-Gym~\citep{jain2025r2egym} and evaluate on SWE-bench
Verified~\citep{jimenez2024swebench} and SWE-bench-Live
Lite~\citep{zhang2026swe}. Training rollouts use a maximum
of 40 agent steps; evaluation uses 50 steps, temperature 1, and three
evaluation runs per checkpoint. We report average accuracy across
these runs and Pass@3, the fraction of tasks solved by at least one
of the three sampled trajectories. In the main BSR setting, we refine
only the failure bucket, with $a_0=0.15$, $a_1=0$, $\delta_0=0.5$,
and $\tau_0=0.5$. This configuration performs similarly to or better
than refining both outcome buckets while requiring fewer pairwise
comparisons (Section~\ref{sec:training-group-ablation}).
The subsequent ablations and trajectory-behavior analysis use
SWE-bench Verified. Complete training settings are provided in Appendix~\ref{app:setup}.

\begin{table}[h]
\centering

\setlength{\tabcolsep}{4pt}
\caption{Performance (\%) on SWE-bench Verified and SWE-bench-Live
Lite. Standard deviations are across
three evaluation runs.}
\label{tab:main-results}
\begin{tabular}{lllcccc}
\toprule
\multicolumn{3}{c}{}
& \multicolumn{2}{c}{SWE-bench Verified}
& \multicolumn{2}{c}{SWE-bench-Live Lite} \\
\cmidrule(lr){4-5}
\cmidrule(lr){6-7}
Size & Budget & Method
& Avg. Acc. $\pm$ Std. & Pass@3
& Avg. Acc. $\pm$ Std. & Pass@3 \\
\midrule

\multirow{4}{*}{14B}
& \multirow{2}{*}{270}
& Baseline
& $31.0 \pm 0.5$ & 45.0
& $7.11 \pm 0.19$ & 13.33 \\
&
& BSR (Ours)
& $\mathbf{33.2 \pm 1.4}$ & \textbf{46.6}
& $\mathbf{8.67 \pm 0.88}$ & \textbf{15.33} \\

\cmidrule(lr){2-7}

& \multirow{2}{*}{400}
& Baseline
& $32.1 \pm 1.1$ & 46.6
& $6.33 \pm 0.34$ & 11.33 \\
&
& BSR (Ours)
& $\mathbf{34.7 \pm 1.0}$ & \textbf{48.0}
& $\mathbf{8.11 \pm 1.02}$ & \textbf{13.67} \\

\midrule

\multirow{2}{*}{32B}
& \multirow{2}{*}{270}
& Baseline
& $34.4 \pm 1.7$ & 48.8
& $7.67 \pm 0.88$ & 12.33 \\
&
& BSR (Ours)
& $\mathbf{35.5 \pm 0.8}$ & \textbf{50.6}
& $\mathbf{8.56 \pm 0.51}$ & \textbf{13.67} \\

\bottomrule
\end{tabular}
\end{table}

% \vspace{-0.2em}
\subsection{Main Results}
\label{sec:main-results}
\begin{figure}[t!]
    \centering

    \begin{subfigure}{0.48\textwidth}
        \centering
        \includegraphics[width=\textwidth]{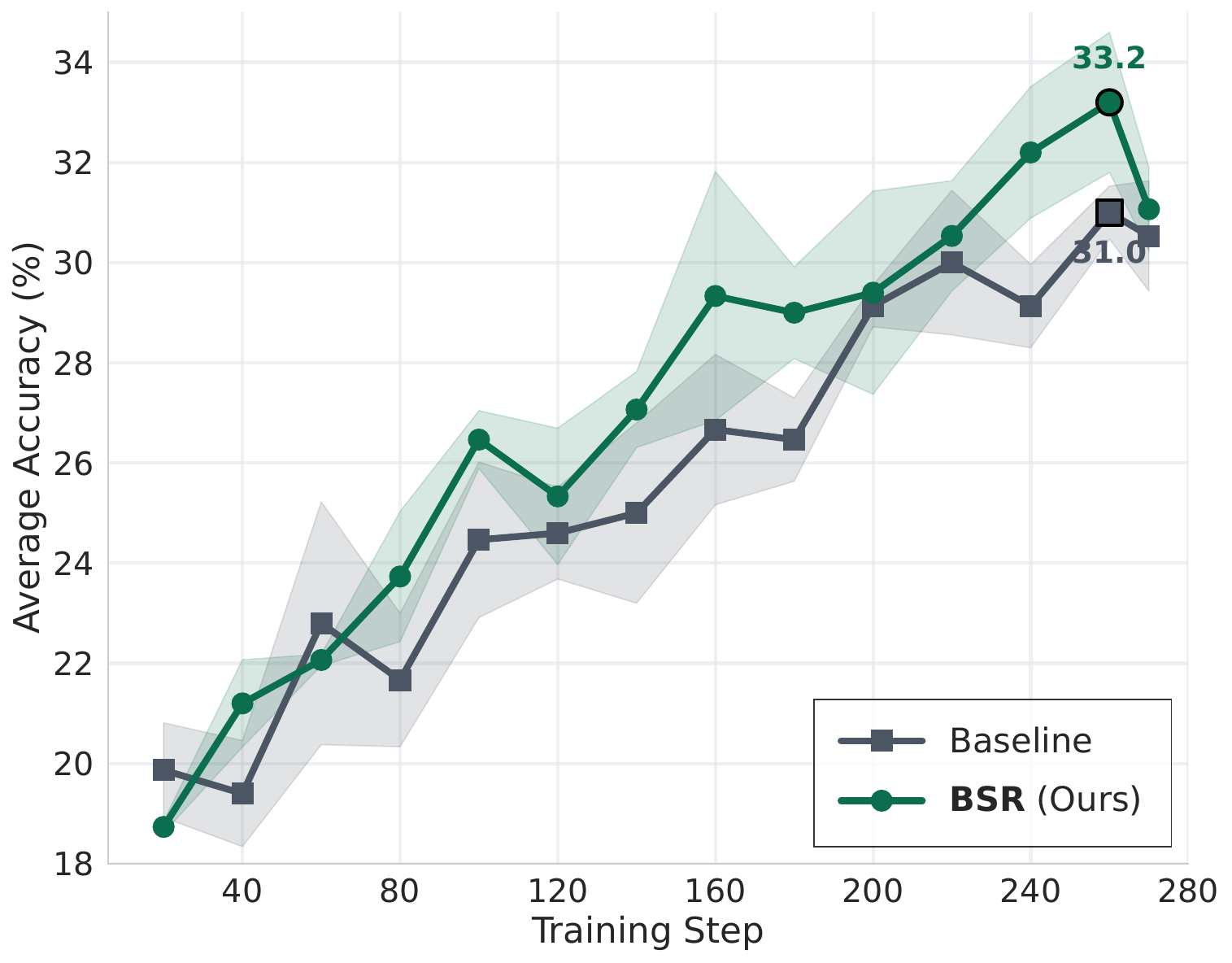}
        \caption{Qwen3-14B}
        \label{fig:14b-avg}
    \end{subfigure}
    \hfill
    \begin{subfigure}{0.48\textwidth}
        \centering
        \includegraphics[width=\textwidth]{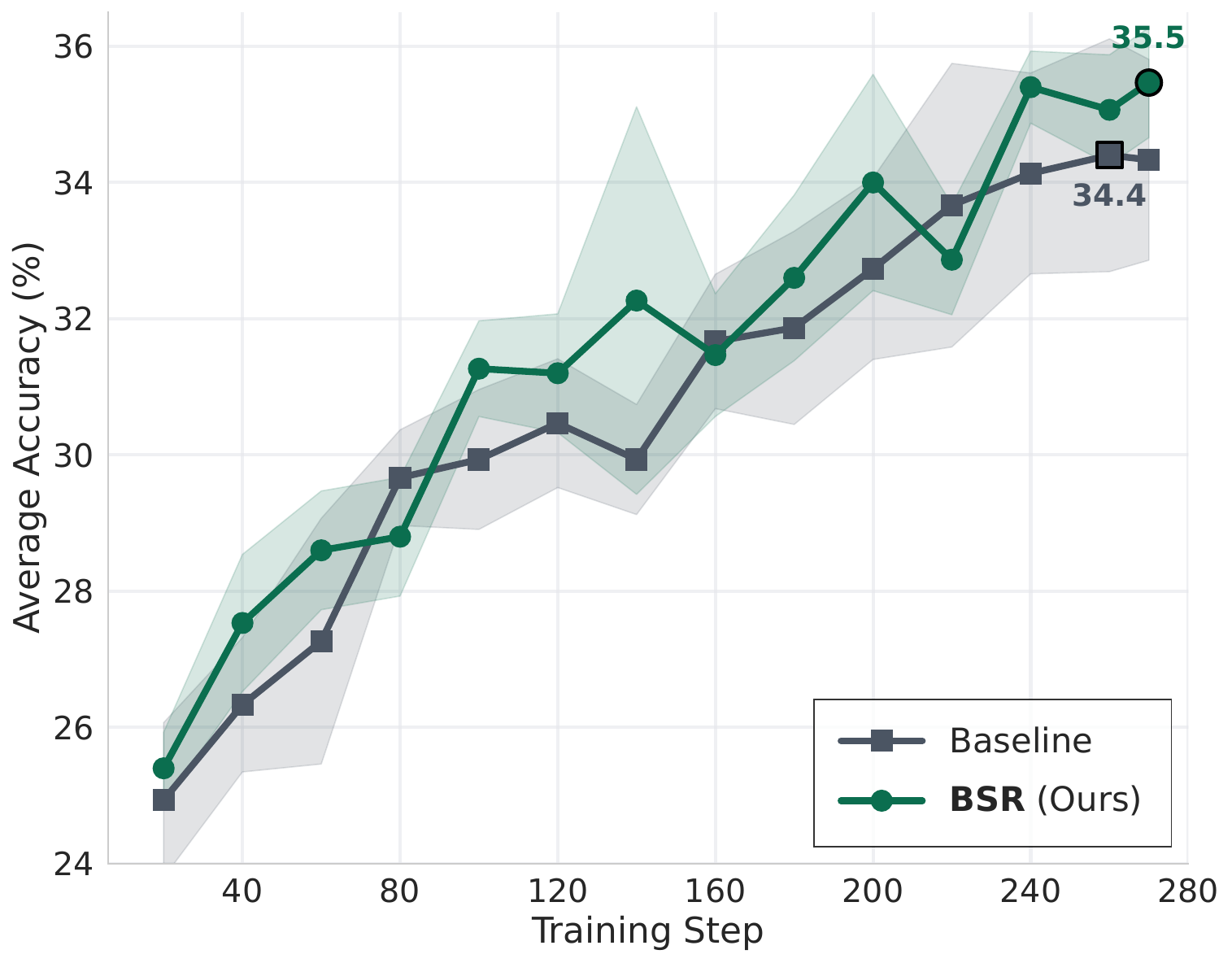}
        \caption{Qwen3-32B}
        \label{fig:32b-avg}
    \end{subfigure}

\caption{SWE-bench Verified evaluation accuracy over training for
Qwen3-14B and Qwen3-32B. Each point averages three evaluation runs.}
    \label{fig:training-curves}
    
\end{figure}

% \vspace{-0.2em}
Table~\ref{tab:main-results} compares BSR with verifier-only RL across
two model sizes, predefined training-step budgets, and two SWE benchmarks.
BSR improves both average accuracy and Pass@3 in every evaluated setting
on SWE-bench Verified and SWE-bench-Live. For example, on
SWE-bench Verified, Qwen3-14B improves from 32.1\% to 34.7\% average
accuracy under the 400-step budget. On SWE-bench-Live, the same
setting improves average accuracy from 6.33\% to 8.11\%. The gains hold
across Qwen3-14B and Qwen3-32B and across both training budgets,
indicating that the improvement is not specific to a single model scale
or training duration. Figure~\ref{fig:training-curves} shows the
corresponding SWE-bench Verified learning curves. We additionally evaluate on Terminal-Bench 2.1~\citep{merrill2026terminal}. BSR improves average
accuracy across all three evaluated model and training-budget settings, full results are provided in Appendix~\ref{app:terminalbench}.

\begin{table}[h!]
\centering
\small
\begin{tabular}{lcc}
\toprule
Statistic & Baseline & BSR (Ours) \\
\midrule
\multicolumn{3}{l}{\textbf{Process metrics}} \\
Related file viewed before first edit $\uparrow$ & 0.733 & \textbf{0.748} \\
Search-to-edit conversion $\uparrow$ & 0.465 & \textbf{0.494} \\
Post-edit test rate $\uparrow$ & 0.301 & \textbf{0.311} \\
Duplicate command rate $\downarrow$ & 0.209 & \textbf{0.208} \\
\midrule
\multicolumn{3}{l}{\textbf{Selected workload statistics}} \\
Average number of errors $\downarrow$ & 13.38 & \textbf{12.76} \\
Average number of searches & 4.74 & 4.06 \\
Average number of tests & 0.747 & 0.797 \\
Average total time (s) $\downarrow$ & 1132.28 & \textbf{1068.55} \\
\bottomrule
\end{tabular}
\caption{Trajectory behavior statistics over the full set of trajectories. Arrows indicate the preferred direction for metrics where applicable. Details are in Appendix~\ref{app:trajectory_behavior_stats}.}
\label{tab}
\end{table}

\paragraph{Behavioral changes.}
Table~\ref{tab} compares investigation, editing,
and verification behavior. BSR more often views a related 
file to the issue before the first edit (0.748 versus 0.733), achieves higher
search-to-edit conversion (0.494 versus 0.465), and performs more
post-edit testing (0.311 versus 0.301). These changes coincide with
fewer searches and recorded errors, alongside a higher average number
of tests. Overall, the aggregate pattern is consistent with stronger
reference-oriented investigation, search follow-through, and
verification. Full trajectory statistics and metric definitions are
provided in Appendix~\ref{app:trajectory_behavior_stats}.

% \yr{Evaluate on SWE-Bench Pro?}

\begin{wraptable}{r}{0.51\textwidth}
\vspace{-1em}
\centering
\small
\begin{tabular}{lcc}
\toprule
Reward mapping & Avg. & Pass@3 \\
\midrule
No deadzone, no softness & 32.0 & 45.6 \\
Deadzone, no softness & 32.6 & 44.0 \\
Deadzone + softness \textbf{(BSR)} & \textbf{33.2} & \textbf{46.6} \\
\bottomrule
\end{tabular}
\caption{Reward-mapping ablation on 14B.}\label{tab:ablation-deadzone-softness}
\vspace{-0.8em}
\end{wraptable}

\subsection{Ablation: Deadzone and Softness}

We examine how the deadzone and softness affect the conversion of
pairwise rankings into auxiliary rewards. The deadzone suppresses
near-zero normalized scores before centering, while softness controls
how deviations outside the deadzone are transformed.
Table~\ref{tab:ablation-deadzone-softness} compares the full mapping
with variants that remove both components or retain the deadzone
without softness. Removing both yields 32.0\% average accuracy and
45.6\% Pass@3. Adding the deadzone alone raises average accuracy to
32.6\%, but lowers Pass@3 to 44.0\%. The full BSR mapping achieves
33.2\% average accuracy, improving both metrics over
either simplified variant. Thus, the deadzone alone does not improve
every metric, whereas combining it with softness gives the strongest
results among the tested mappings. Full training curves are provided
in Figure~\ref{fig:deadzone-ablation} in appendix.

\begin{figure}[t]
\centering

\begin{minipage}[t]{0.48\textwidth}
    \centering
    \vspace{0pt}
    \small
    \begin{tabular}{lcc}
    \toprule
    Refined bucket(s) & Avg. & Pass@3 \\
    \midrule
    Baseline & 31.0 & 45.0 \\
    \midrule
    Pass bucket only & 32.0 & 44.0 \\
    Pass + fail buckets & 33.1 & 44.6 \\
    Fail bucket only \textbf{(BSR)} & \textbf{33.2} & \textbf{46.6} \\
    \bottomrule
    \end{tabular}
    \captionof{table}{Ablation of the verifier-outcome buckets receiving BSR (\%).}
    \label{tab:ablation-training-group}
\end{minipage}
\hfill
\begin{minipage}[t]{0.48\textwidth}
    \centering
    \vspace{0pt}
    \small
    \begin{tabular}{lcc}
    \toprule
    Method & Avg. acc. & Pass@3 \\
    \midrule
    Baseline & 31.0 & 45.0 \\
    Random reward & 30.9 & 42.8 \\
    BSR \textbf{(Ours)} & \textbf{33.2} & \textbf{46.6} \\
    \bottomrule
    \end{tabular}
    \captionof{table}{
    Random-reward ablation on Qwen3-14B. Random rewards use the same amplitude range as BSR
    but ignore behavioral rankings.
    }
    \label{tab:random-reward-ablation}

\vspace{-0.5em}
\end{minipage}
% \hfill

\end{figure}

\subsection{Ablation: Verifier-Outcome Buckets}
\label{sec:training-group-ablation}

We compare applying BSR to the failure bucket, the success bucket,
or both. The failure bucket includes failed trajectories from both
all-fail and mixed-outcome rollout groups.
Table~\ref{tab:ablation-training-group} shows that failure-only
refinement achieves the highest average accuracy and Pass@3, at
33.2\% and 46.6\%, respectively. Refining both buckets achieves
similar average accuracy of 33.1\%, but lower Pass@3 of 44.6\%,
while requiring additional comparisons among successful trajectories.
Success-only refinement reaches 32.0\% average accuracy and
44.0\% Pass@3: it improves average accuracy over the verifier-only
baseline, but does not improve coverage across three sampled solutions.
These results favor failure-only refinement in this setting.
Extending refinement to successful trajectories increases judgment
workload without improving either reported metric over failure-only
refinement. Full training curves appear in
Figure~\ref{fig:group-ablation} in the appendix.

% \subsection{Analysis: Agreement with an External Judge}

% We examine how the policy's pairwise judgments agree with a strong external
% judge as RL training progresses. Every 10 training steps, we randomly
% sample 200 pairwise trajectory comparisons from Qwen3-14B training
% and re-evaluate the same pairs using Kimi K2.5~\citep{team2026kimi} as a reference judge.
% Figure~\ref{fig:judgment-alignment} reports macro F1, which is generally
% higher at early checkpoints and decreases over training. Since new
% on-policy pairs are sampled at each checkpoint, this measures agreement
% on an evolving comparison distribution rather than isolating changes
% in the judge on a fixed set of trajectories. Kimi K2.5 provides
% a proxy reference rather than ground-truth labels.
% We use this analysis to characterize the agreement and
% limitations of the self-judge, rather than as an absolute measure of
% judgment quality. Despite the observed decline, the main
% results show improved performance with BSR while reusing the
% same model for reward refinement, without a separate
% frozen judge.

\subsection{Additional Study: Random-reward control}

We test whether arbitrary auxiliary rewards can reproduce BSR's gains
without behavioral rankings. The control assigns random rewards within
the same amplitude range as BSR, but does not associate them with
pairwise trajectory preferences.
Table~\ref{tab:random-reward-ablation} shows that the control achieves
30.9\% average accuracy and 42.8\% Pass@3, compared with 31.0\% and
45.0\% for verifier-only RL. BSR instead reaches 33.2\% average
accuracy and 46.6\% Pass@3. Random rewards therefore fail to improve
average accuracy and reduce Pass@3, whereas BSR improves both.
This shows that matching the reward amplitude alone
is insufficient to reproduce the gains from BSR's ranking-based
refinement.

%% file: appendix.tex
\section{Additional Method Details}
\subsection{Pairwise Comparison Scheduling.}
Algorithm~\ref{alg:pairwise-compare} describes the comparison schedule for each
verifier-outcome bucket $B_y$, where $y\in\{0,1\}$ denotes the shared verifier
outcome. For $2\leq |B_y|<5$, we use exhaustive pairwise comparison. For larger
buckets, we use a Swiss-style tournament schedule~\citep{sauer2024improving}
to reduce the number of model calls while maintaining balanced comparison
coverage. Table~\ref{tab:pairwise-schedule-reduction} shows the resulting
reduction in model comparisons.

For each scheduled pair, we ask the model to compare the two trajectory
summaries given the task $x$, reference patch $p^\star$, and bucket outcome
$y$. The model selects the better trajectory under the shared outcome: better
progress among failures or cleaner behavior among successes. We denote the
collection of performed comparisons by $\mathcal{C}_y$, where each comparison
record $(i,j,w_{ij},\alpha_{ij})$ contains the compared pair $(i,j)$, the
preferred trajectory $w_{ij}$, and the preference-strength label
$\alpha_{ij}$. To reduce position bias during BSR training, we randomize the
presentation order of each pairwise comparison.

We use separate prompts for the pass and fail buckets
(Figures~\ref{prompt:success} and~\ref{prompt:fail}). They have the same
general structure but direct the model to different behavioral criteria
depending on the verifier outcome. We show the user prompt in
Figure~\ref{prompt:user} and an example ranking response in
Figure~\ref{example_compare}.

\algrenewcommand{\algorithmiccomment}[1]{\hfill{\footnotesize$\triangleright$ #1}}

\begin{algorithm}[h]
\caption{\textsc{PairwiseCompare}: Comparison Scheduling for One Outcome Bucket}
\label{alg:pairwise-compare}
\small
\begin{algorithmic}[1]
\Require Task $x$, reference patch $p^\star$, model $\pi_\theta$,
trajectory summaries $\{h_i:i\in B_y\}$, bucket $B_y$, outcome $y\in\{0,1\}$
\State $n\gets |B_y|$
\State $\mathcal{C}_y\gets \emptyset$
\If{$n<2$}
    \State \Return $\mathcal{C}_y$
    \Comment{No pairwise comparison is possible}
\ElsIf{$n<5$}
    \Comment{\textbf{Case 1: Exhaustive comparison}}
    \State $\mathcal{P}\gets\{(i,j):i,j\in B_y,\ i<j\}$
    \For{$(i,j)\in\mathcal{P}$}
        \State $(w_{ij},\alpha_{ij})
        \gets\textsc{ModelCompare}_{\pi_\theta}(x,p^\star,h_i,h_j,y)$
        \State Append $(i,j,w_{ij},\alpha_{ij})$ to $\mathcal{C}_y$
    \EndFor
\Else
    \Comment{\textbf{Case 2: Swiss-style comparison}}
    \State $T\gets\min(\lceil\log_2 n\rceil+2,\ n-1)$
    \Comment{Number of rounds}
    \State Initialize wins $v_i\gets0$ and match counts $c_i\gets0$
    for all $i\in B_y$
    \State Initialize pair counts $c_{ij}\gets0$ for all $i,j\in B_y$
    \For{$t=1,\ldots,T$}
        \State Sort trajectories by decreasing $v_i$, then increasing $c_i$
        \State $\mathcal{P}_t\gets\emptyset$
        \State Form round pairs greedily among unused trajectories
        \Comment{Prefer unseen pairs, fewer rematches, and similar win counts}
        \State Update $c_{ij}$ for each scheduled pair
        $(i,j)\in\mathcal{P}_t$
        \For{$(i,j)\in\mathcal{P}_t$}
            \State $(w_{ij},\alpha_{ij})
            \gets\textsc{ModelCompare}_{\pi_\theta}(x,p^\star,h_i,h_j,y)$
            \State Append $(i,j,w_{ij},\alpha_{ij})$ to $\mathcal{C}_y$
            \State $v_{w_{ij}}\gets v_{w_{ij}}+1$
            \Comment{Update Swiss score}
            \State $c_i\gets c_i+1,\quad c_j\gets c_j+1$
            \Comment{Update match counts}
        \EndFor
    \EndFor
\EndIf
\State \Return $\mathcal{C}_y$
\end{algorithmic}
\end{algorithm}

\begin{table}[h]
\centering
\small
\begin{tabular}{c c c c}
\toprule
Bucket size $n$ & Full comparisons $\binom{n}{2}$ &
Scheduled comparisons & Reduction \\
\midrule
1 & 0  & 0  & -- \\
2 & 1  & 1  & 0.0\% \\
3 & 3  & 3  & 0.0\% \\
4 & 6  & 6  & 0.0\% \\
5 & 10 & 8  & 20.0\% \\
6 & 15 & 15 & 0.0\% \\
7 & 21 & 15 & 28.6\% \\
8 & 28 & 20 & 28.6\% \\
\bottomrule
\end{tabular}
\caption{Comparison reduction from the scheduling scheme. For $n<5$, the
scheduler uses exhaustive pairwise comparison. For $n\geq5$, it uses
$T=\min(\lceil\log_2 n\rceil+2,n-1)$ Swiss-style rounds, with at most
$\lfloor n/2\rfloor$ comparisons per round.}
\label{tab:pairwise-schedule-reduction}
\end{table}

\subsection{Weighted Bradley--Terry Score Fitting}
\label{app:weighted-btl}

\paragraph{Weighted Bradley--Terry aggregation.}
Given the performed comparisons $\mathcal{C}_y$ for an outcome bucket $B_y$,
we convert pairwise preferences into task-local scalar scores using a weighted
Bradley--Terry model~\citep{bradley1952rank}. Each trajectory $i\in B_y$ is
assigned a latent score $s_i$. For a comparison between trajectories $i$ and
$j$, the model defines
\begin{equation}
P(i\succ j)
=
\sigma(s_i-s_j),
\qquad
\sigma(t)=\frac{1}{1+\exp(-t)}.
\end{equation}
Let $m_{ij}=1$ if $w_{ij}=i$ and $m_{ij}=0$ otherwise. We fit the scores by
minimizing the weighted negative log likelihood with $\ell_2$ regularization:
\begin{equation}
\mathcal{L}_{\mathrm{BTL}}(s)
=
-
\sum_{(i,j,w_{ij},\alpha_{ij})\in\mathcal{C}_y}
\alpha_{ij}
\left[
m_{ij}\log\sigma(s_i-s_j)
+
(1-m_{ij})\log\sigma(s_j-s_i)
\right]
+
\frac{\lambda}{2}\sum_{i\in B_y}s_i^2.
\end{equation}
Algorithm~\ref{alg:btl-fit} gives the optimization procedure used to obtain
the scores. We use a diagonal second-order preconditioner and re-center the
scores after each iteration, since the downstream reward map depends only on
their relative values.

\begin{algorithm}[th]
\caption{Weighted BTL Score Fitting}
\label{alg:btl-fit}
\small
\begin{algorithmic}[1]
\Require Bucket $B_y$, comparisons $\mathcal{C}_y$,
regularization $\lambda$, learning rate $\eta$
\State Initialize $s_i\gets0$ for all $i\in B_y$
\Repeat
    \For{$i\in B_y$}
        \State $g_i\gets\lambda s_i$
        \State $d_i\gets\lambda$
    \EndFor
    \For{$(i,j,w_{ij},\alpha_{ij})\in\mathcal{C}_y$}
        \State $m_{ij}\gets\mathbf{1}\{w_{ij}=i\}$
        \State $p_{ij}\gets\sigma(s_i-s_j)$
        \State $g_i\gets g_i+\alpha_{ij}(p_{ij}-m_{ij})$
        \State $g_j\gets g_j-\alpha_{ij}(p_{ij}-m_{ij})$
        \State $d_i\gets d_i+\alpha_{ij}p_{ij}(1-p_{ij})$
        \State $d_j\gets d_j+\alpha_{ij}p_{ij}(1-p_{ij})$
    \EndFor
    \For{$i\in B_y$}
        \State $s_i\gets s_i-\eta\,g_i/(d_i+\epsilon)$
    \EndFor
    \State Center scores:
    $s_i\gets s_i-\frac{1}{|B_y|}\sum_{j\in B_y}s_j$
\Until{convergence}
\State \Return scores $\{s_i:i\in B_y\}$
\end{algorithmic}
\end{algorithm}

\subsection{Trajectory Summarization}
\label{app:trajectory-summarization}

Before pairwise judgment, we convert each raw agent trajectory into a compact
step-wise representation $h_i$. We parse the interaction transcript into an
ordered sequence of tool actions and resulting observations. Because the judge
is intended to compare externally visible debugging behavior, we remove the
model's internal thinking and retain the action-observation trace. We use
deterministic reductions whenever possible and only fall back to LLM-based
summarization when an observation remains too long to preserve directly.

\paragraph{Action reparsing.}
Tool actions are originally emitted in a verbose function-call format. We parse
each assistant message to extract the tool name and its parameters, and render
the action as a compact function-style call, such as
\texttt{tool\_name(arg=value)}. This preserves the invoked tool and argument
values while removing formatting overhead such as XML-style wrappers.
Multiline values are kept as multiline strings, structured values such as lists
or dictionaries are preserved, numbers and booleans are left unquoted, and
ordinary strings are JSON-escaped. If no valid tool call is found, we retain
the original action text; if a tool is found but has no parameters, we render
it as a zero-argument call.

\paragraph{Observation reduction.}
Observations are reduced using tool-specific deterministic rules before any
LLM-based summarization. Short observations are kept unchanged. For bash
outputs, observations containing errors, tracebacks, or exceptions are
preserved so that failure information remains visible to the judge. For
file-editor observations, we distinguish file views from other editor
operations. When a long file view is produced by a numbered
\texttt{cat}-style output, we retain the first and last portions and insert a
marker indicating that the middle was omitted. Other compact file-editor
outputs are kept unchanged. If an observation remains long and does not match
one of these safe reduction cases, we summarize it with an
observation-specific prompt that preserves the key command outcome, relevant
files or paths, and any errors.

\subsection{Other Potential Instantiations of Behavioral Reward}
\label{app:alternative-behavioral-rewards}

Beyond pairwise ranking, we explored two alternative instantiations of
behavioral reward: group ranking and rubric grading. In preliminary experiments,
neither produced a clear improvement, so we did not tune or investigate them
further. Nevertheless, they illustrate other ways in which behavioral judgment
could be used to refine verifier rewards.

\paragraph{Group ranking.}
Group ranking evaluates all trajectories in the same verifier-outcome bucket
jointly. Each trajectory is first summarized through a structured process
evaluation covering aspects such as localization, root-cause analysis, fix
quality, verification, and overall behavior. The judge then considers the full
group and assigns each trajectory a coarse score from 1 to 5, with ties
allowed. These scores are mapped to bounded auxiliary rewards within the
corresponding verifier-outcome bucket. Compared with pairwise ranking, this
approach gives the judge a global view of the group and can reduce the number
of direct comparison calls.

\paragraph{Rubric grading.}
Rubric grading evaluates each trajectory independently against a predefined set
of behavioral criteria. The judge determines whether each progress criterion
is satisfied and whether any undesirable behavior, represented as a red flag,
is present. Unsatisfied progress criteria and triggered red flags produce
weighted penalties that refine the verifier reward. This approach is simple,
parallelizable, and provides an absolute behavioral assessment without
requiring comparisons among trajectories.

\paragraph{Discussion.}
The three approaches make different trade-offs. Pairwise ranking provides
direct relative supervision and requires only local comparisons, but requires
additional judgment compute to evaluate multiple pairs. Group ranking compares
the entire bucket more efficiently and captures global relative structure, but
its scores remain relative to the particular group and may be sensitive to
group composition or listwise presentation. Rubric grading is less expensive
and produces an absolute assessment for each trajectory, but it does not
directly compare trajectories and depends on the completeness and calibration
of the predefined rubric. Conversely, both pairwise and group ranking lack a
task-independent absolute quality scale. Combining absolute rubric-based
assessment with comparative ranking may therefore be a useful direction for
future work.

\begin{table}[t]
\centering
\begin{tabular}{|l|c|}
\hline
\textbf{Parameter} & \textbf{Value} \\
\hline

\multicolumn{2}{|c|}{\textbf{Model Configuration}} \\
\hline
Max Prompt Length & 4096 \\
Max Response Length & 32768 \\
\hline

\multicolumn{2}{|c|}{\textbf{Training Settings}} \\
\hline
Train Dataset & R2E-Gym Subset \\
Train Batch Size & 64 \\
Learning Rate & $1\times10^{-6}$ \\
\hline

\multicolumn{2}{|c|}{\textbf{RL Settings}} \\
\hline
Advantage Estimator & LOOP \\
Rollout Samples & 8 \\
Rollout Temperature & 1.0 \\
PPO Mini-batch Size & 64 \\
KL Loss & False \\
Entropy Coefficient & 0.0 \\
Clip Ratio High & 0.28 \\
Training Max Agent Steps & 40 \\
\hline

\multicolumn{2}{|c|}{\textbf{BSR Reward Settings}} \\
\hline
Rewarded Outcome Bucket & Fail ($y=0$) \\
Reward Amplitude $a_0$ & 0.15 \\
Deadzone Threshold $\delta_0$ & 0.5 \\
Softness $\tau_0$ & 0.5 \\
\hline

\end{tabular}
\caption{Important hyperparameters used in the main experiments.}
\label{tab:swe_agent_ppo_hyperparams}
\end{table}

\section{Additional Experiment Details, Results, and Analysis}
\subsection{Detailed Setup}
\label{app:setup}

We follow RLLM for training and use the filtered R2E-Gym dataset
(4,578 tasks) as our training set. We show important parameters in Table~\ref{tab:swe_agent_ppo_hyperparams}. We evaluate on SWE-bench Verified
(500 tasks) and SWE-bench-Live Lite (300 tasks). SWE-bench Verified is evaluated
periodically throughout training. For both benchmarks, evaluation uses
a maximum of 50 agent steps, temperature 1, and three runs. For each benchmark, we report the mean and standard deviation of
accuracy across three evaluation runs. Pass@3 is the fraction of
tasks solved in at least one of the three runs.
% Different from RLLM, we change the max steps of SWE agent to 40 steps, because we find that more than 95\% of success trajectories finish in less than 40 steps, and also to speed up training.

For Qwen3-14B, we train on 8 GPU nodes, each containing 8 A100-SXM4-40G GPUs. For Qwen3-32B, we use 16 nodes. Each training run takes 6 to 8 days.

% \paragraph{Question: How does the method perform with different model sizes?}
\paragraph{Training observations.}
For Qwen3-14B, step-limit terminations become more frequent during
training, while the truncation rate initially decreases and later rises.
The logged entropy loss also increases. We observe that more than 95\%
of successful trajectories finish in fewer than 40 steps, motivating
a reduction of the training rollout limit from 50 to 40 steps. In our
experiments, this change improves performance and reduces training time,
whereas lowering the sampling temperature or adding KL regularization
reduces performance. Evaluation retains a maximum of 50 steps.

\paragraph{Kubernetes setting.}
Environment provisioning is a major systems challenge for SWE-agent training, since each optimization step may require 512 concurrent environments ($64$ tasks $\times$ $8$ rollouts). In our experiments, we manually configured a two-node Kubernetes cluster, with each node providing 128 CPU cores and approximately 500\,GB of memory. To reduce startup latency, we pre-pull and cache all required container images. We also use small per-pod resource requests (80\,m CPU and 300\,MiB memory), allowing the cluster to schedule a large number
of concurrent environments. Several configuration details are important in practice:
\begin{itemize}[leftmargin=*]
    \item Ensure that pods have outbound network access when required by the task.
    \item Increase the maximum number of pods allowed per node to support the desired level of parallelism.
    \item Configure Kubernetes and containerd image-garbage-collection and disk-eviction policies to avoid removing frequently used cached images.
    \item Maintain a stable and automatically recoverable connection between the GPU training machines and CPU environment nodes.
    \item Promptly clean up completed or failed pods to prevent resource leakage.
\end{itemize}
Additional setup instructions and configuration examples are included in our codebase.

\paragraph{Performance optimization.}
To reduce end-to-end training time, we overlap environment provisioning with other parts of the training pipeline. First, while the model is being updated for the current batch, we asynchronously pre-create the environments required by the next batch. This hides a substantial portion of environment startup latency behind model optimization. Second, we remove the synchronization barrier between environment creation and trajectory generation. Rather than waiting for all environments in a batch to become ready, each rollout begins as soon as its own environment has been created. This allows fast-starting environments to enter trajectory generation immediately and reduces idle time caused by slow environment initialization or other stragglers.

\paragraph{Training-time overhead.}
\label{app:bsr-overhead}
BSR reuses the policy-serving infrastructure for post-rollout
comparisons, without a separately deployed reward model. In the
measured Qwen3-14B configuration, BSR required 19.86\% more wall-clock
training time than verifier-only RL under the same training and
serving GPU allocation. This additional judging is performed during
training; evaluation uses the acting policy without BSR judge calls.
While our method results in overhead, however our method is extremely easy to apply without additional computation resources and setup. Also, extending verifier-only training
from 270 to 400 steps does not recover the BSR gain: BSR at 270 steps
already exceeds the 400-step baseline in average accuracy. This
suggests that the improvement is not explained simply by additional
optimization steps, although a fully compute-matched comparison remains
an interesting direction. We note that more selective comparison scheduling, improved
batching, and greater overlap between ranking and other training stages
could further reduce the cost.

% \paragraph{Training-time overhead.}
% BSR reuses the same model-serving infrastructure for a lightweight
% post-rollout ranking stage, requiring no additional GPU machines or separately
% deployed reward model or judge service. In our Qwen3-14B experiments, this
% results in 19.86\% additional wall-clock training time under the same
% training and serving GPU allocation. This modest overhead is measured with our
% current implementation, which is not optimized for ranking throughput. More
% selective comparison scheduling, improved batching, and greater overlap between
% ranking and other training stages could further reduce the cost.

\begin{table}[ht]
\centering
\small
\caption{Terminal-Bench 2.1 performance (\%) over 89 tasks and three
evaluation trials.}
\label{tab:terminalbench-results}
\begin{tabular}{lllcc}
\toprule
Size & Budget & Method & Avg. Acc. & Pass@3 \\
\midrule
\multirow{4}{*}{14B}
& \multirow{2}{*}{270}
& Baseline   & 7.12 & 14.61 \\
&
& BSR (ours) & \textbf{7.87} & \textbf{15.73} \\
\cmidrule(lr){2-5}
& \multirow{2}{*}{400}
& Baseline   & 6.74 & 14.61\\
&
& BSR (ours) & \textbf{8.24} & \textbf{15.73} \\
\midrule
\multirow{2}{*}{32B}
& \multirow{2}{*}{270}
& Baseline   & 4.87 & 8.99 \\
&
& BSR (ours) & \textbf{5.62} & \textbf{13.48} \\
\bottomrule
\end{tabular}
\end{table}

\begin{figure}[htbp]
    \centering

    \begin{subfigure}[t]{0.49\linewidth}
        \centering
        \includegraphics[width=\linewidth]{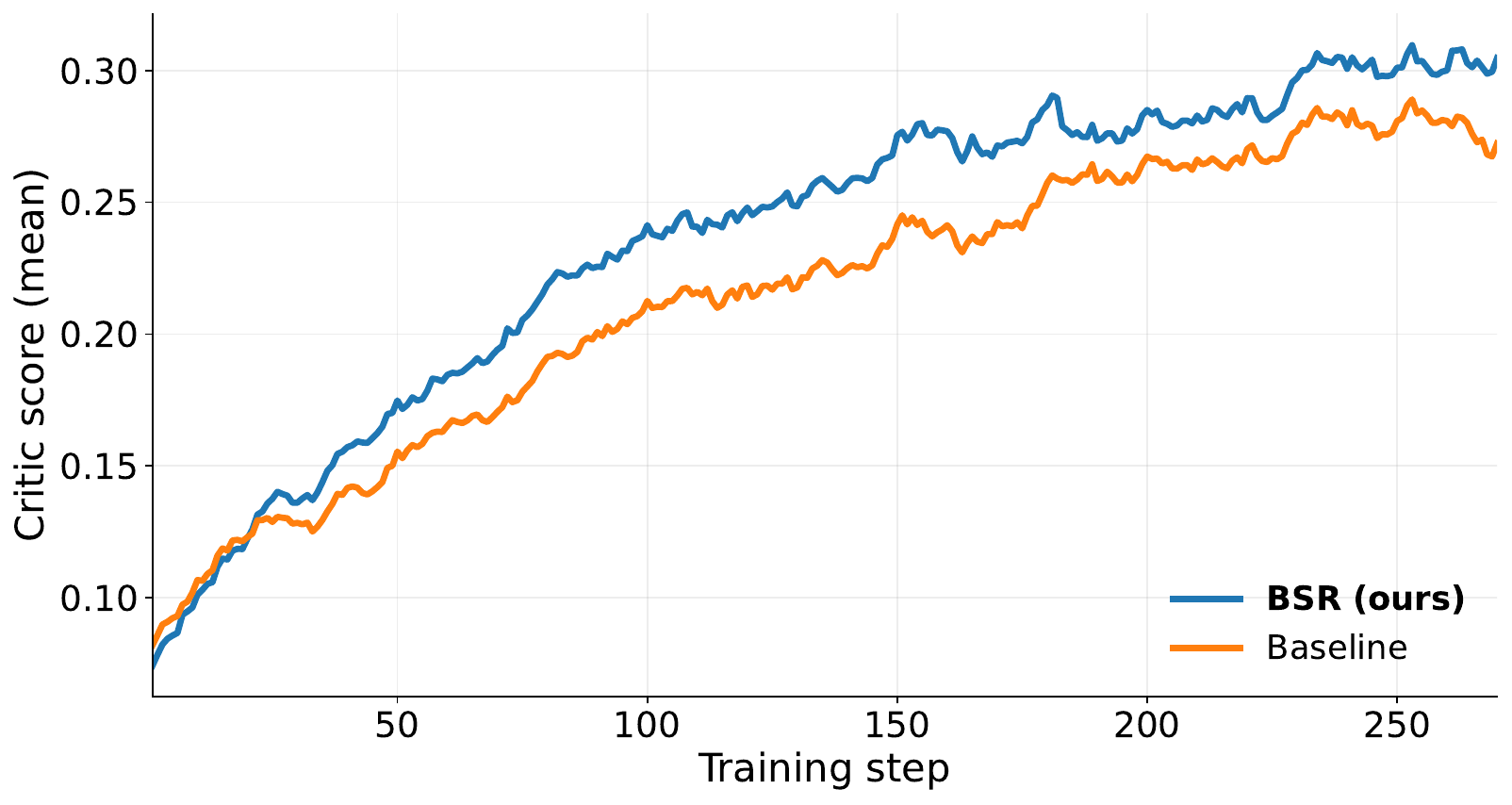}
        \caption{14B model.}
        \label{fig:critic-score-14b}
    \end{subfigure}
    \hfill
    \begin{subfigure}[t]{0.49\linewidth}
        \centering
        \includegraphics[width=\linewidth]{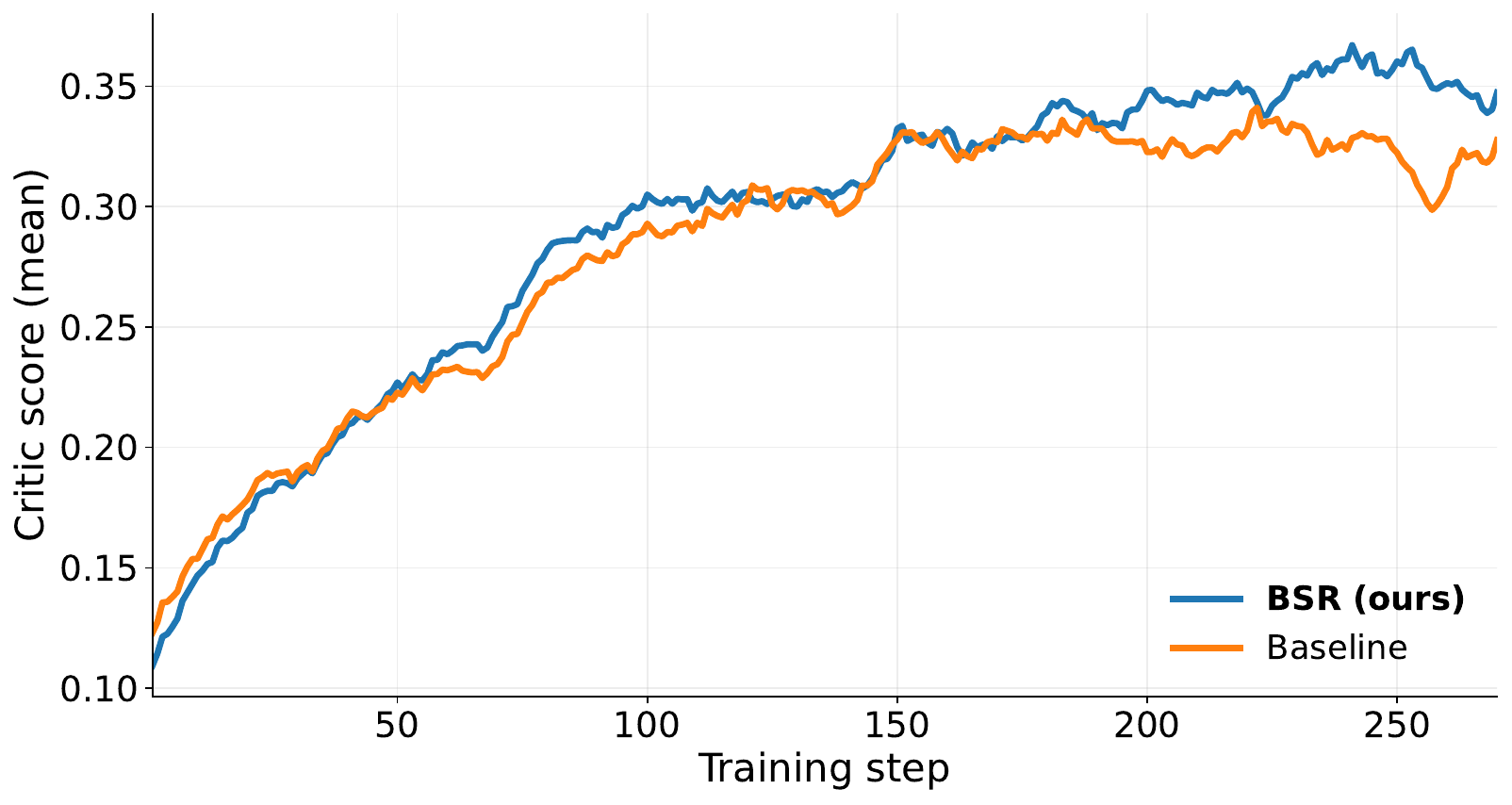}
        \caption{32B model.}
        \label{fig:critic-score-32b}
    \end{subfigure}

    \caption{Mean critic scores on training set for BSR and the baseline. Curves are smoothed with EMA with a span of 31 (Note here the critic score is just the score from the execution verifier).}
    \label{fig:critic-score}
\end{figure}

\begin{figure}[htbp]
    \centering

    \begin{subfigure}{0.85\textwidth}
        \centering
        \includegraphics[width=\textwidth]{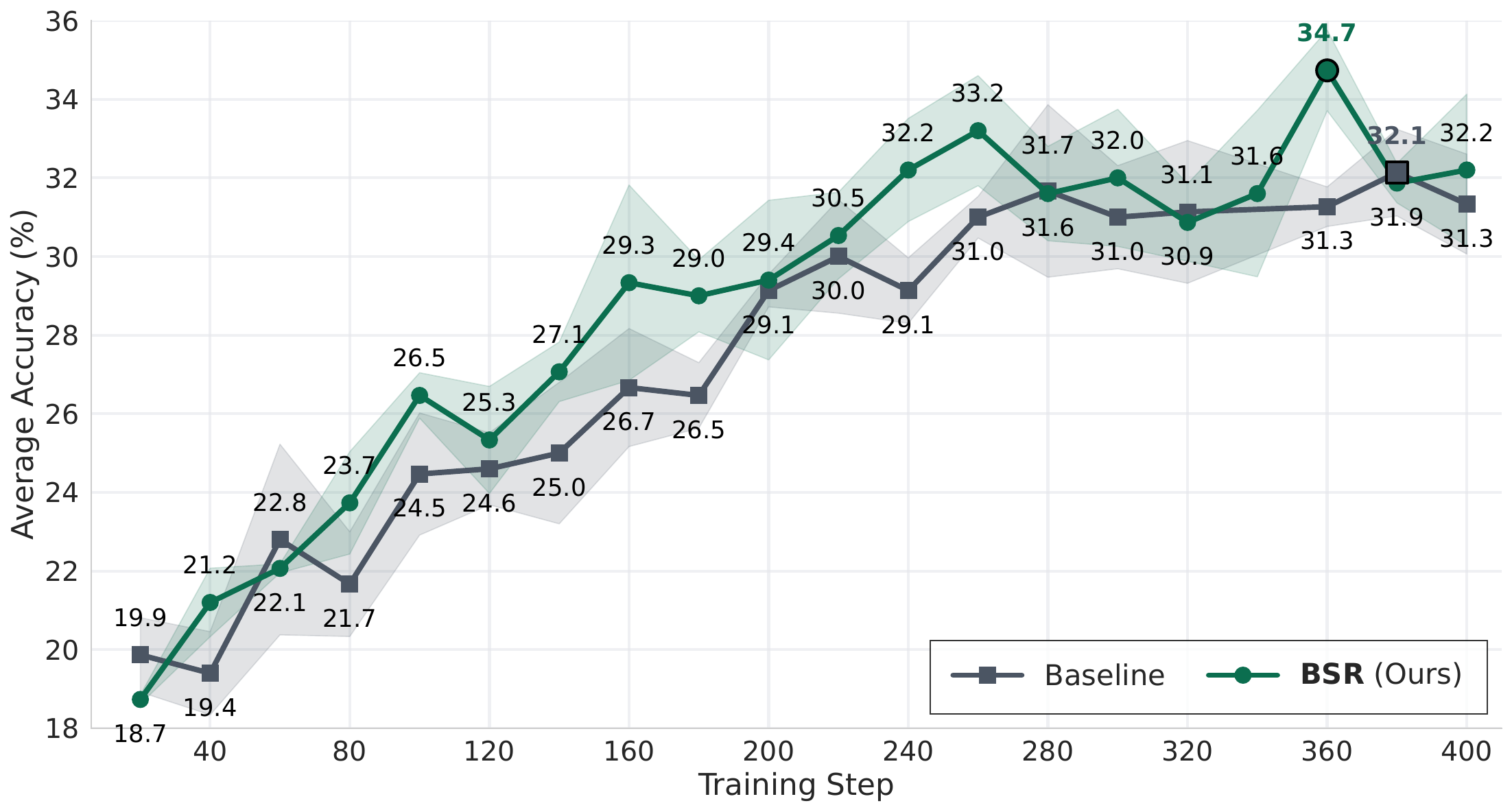}
        \caption{Average Accuracy.}
        \label{fig:14b-full}
    \end{subfigure}

    \vspace{0.5em}

    \begin{subfigure}{0.85\textwidth}
        \centering
        \includegraphics[width=\textwidth]{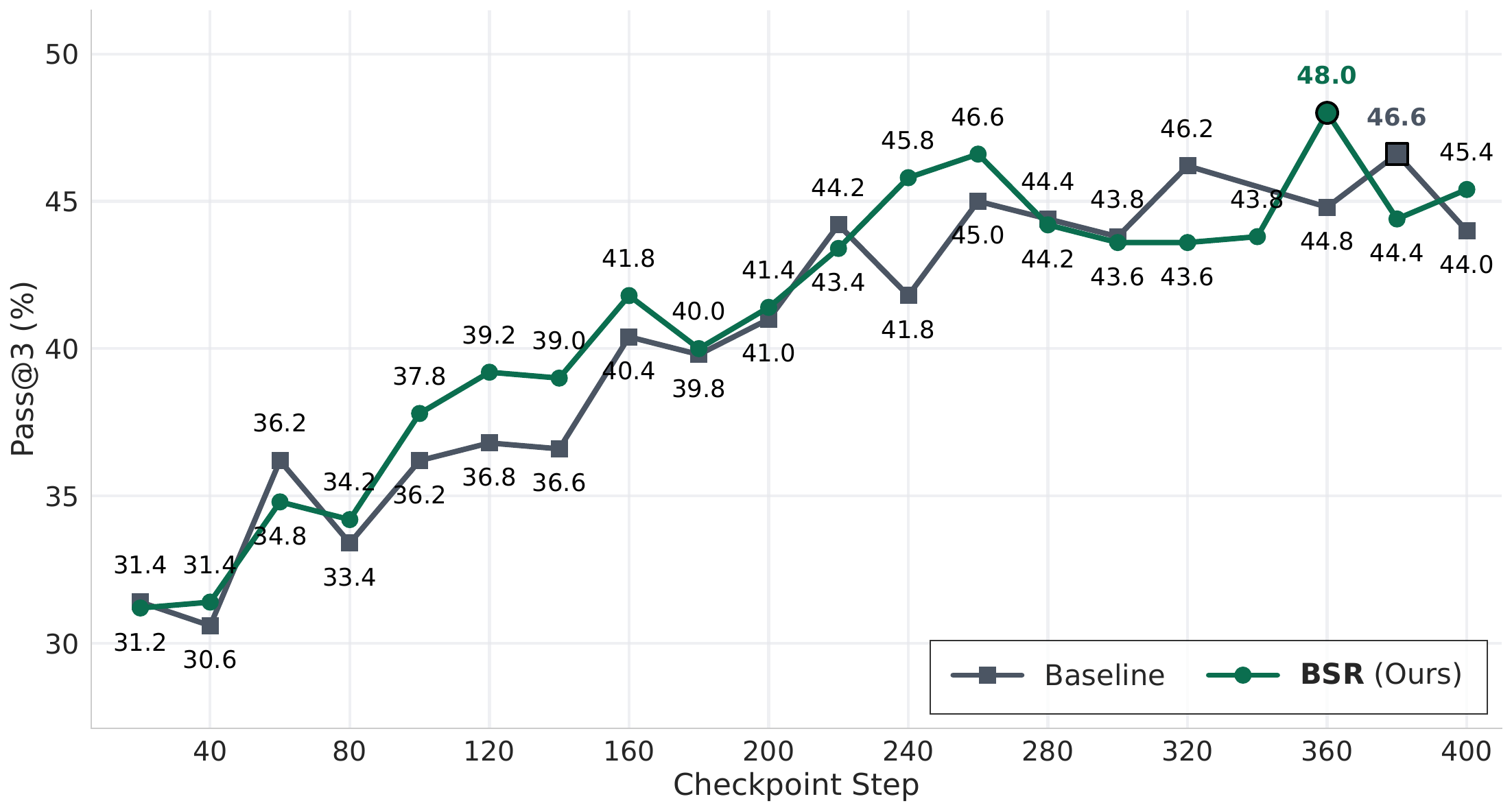}
        \caption{Pass@3.}
        \label{fig:14b-pass-at-3}
    \end{subfigure}

    \caption{Evaluation on SWE-bench Verified while training Qwen3-14B for long run of 400 steps. We evaluate with temperature = 1 and max\_steps = 50 for 3 runs.}
    \label{fig:14b-checkpoint-comparison}
\end{figure}

\begin{figure}[htbp]
    \centering

    \begin{subfigure}{0.48\textwidth}
        \centering
        \includegraphics[width=\textwidth]{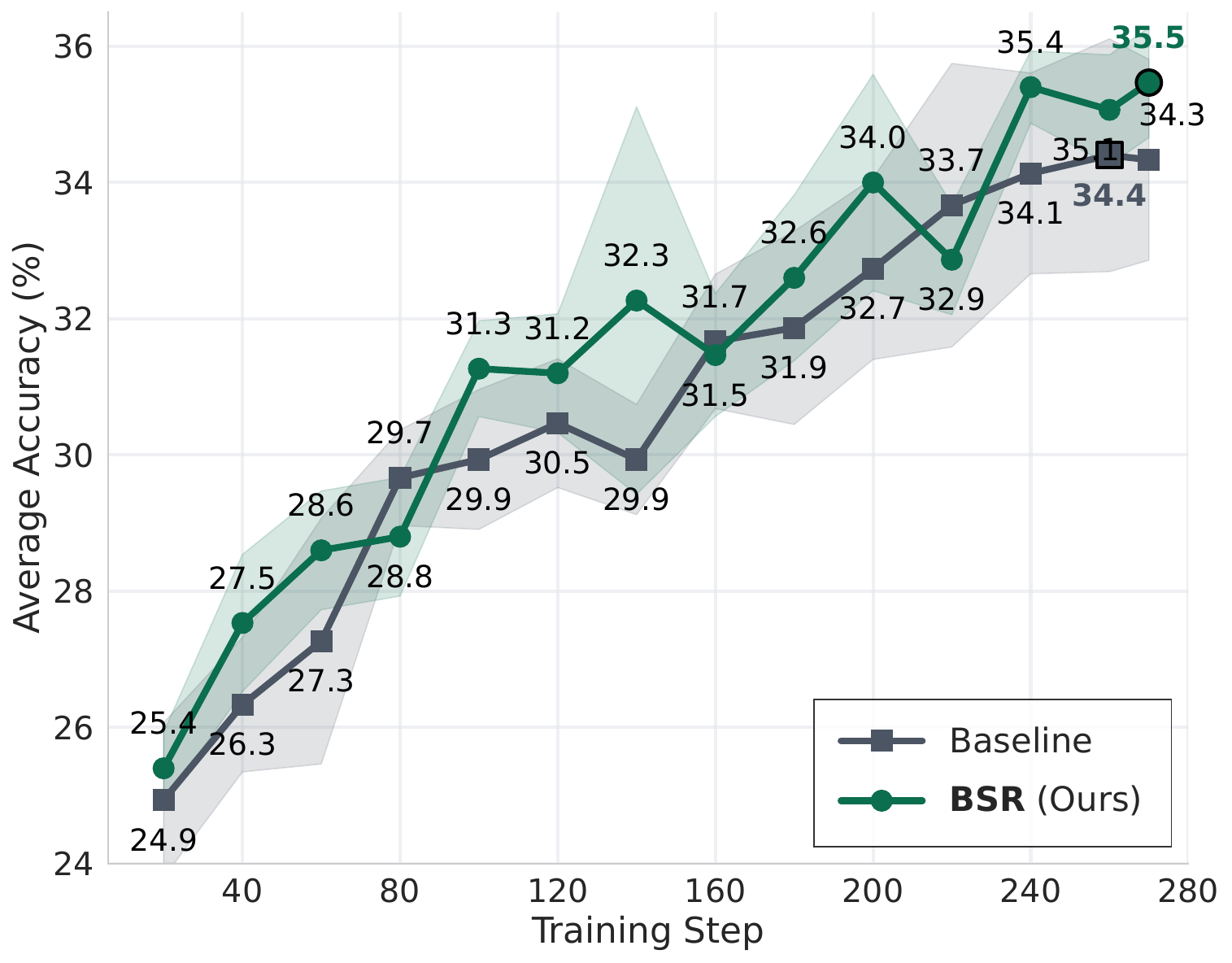}
        \caption{Average Accuracy.}
    \end{subfigure}
    \hfill
    \begin{subfigure}{0.48\textwidth}
        \centering
        \includegraphics[width=\textwidth]{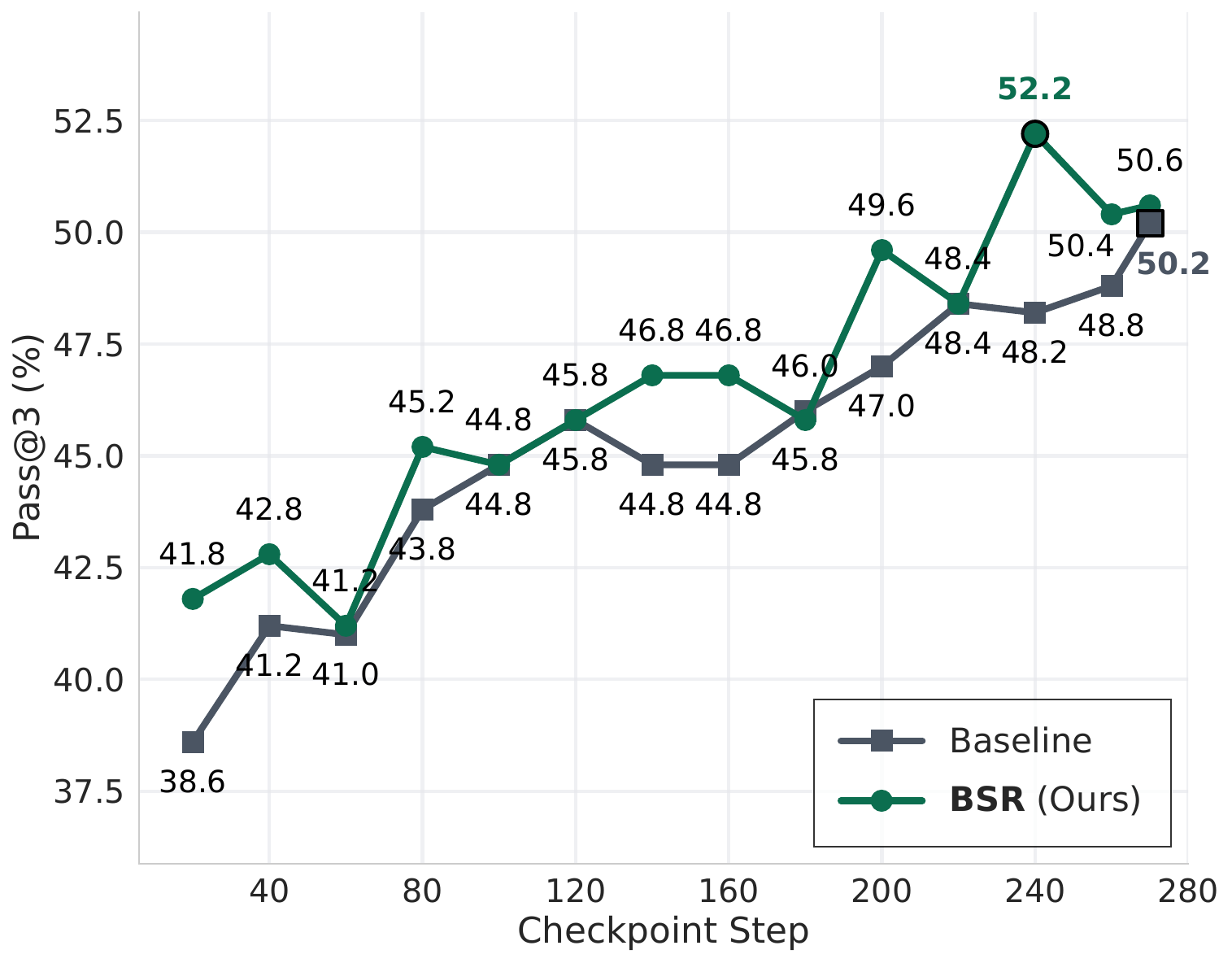}
        \caption{Pass@3.}
    \end{subfigure}

    \caption{Detailed plot of evaluation on SWE-bench Verified while training Qwen3-32B. We evaluate with temperature = 1 and max\_steps = 50 for 3 runs.}
    \label{fig:32b_labeled_all}
\end{figure}

\begin{figure}[htbp]
    \centering

    \begin{subfigure}{0.48\textwidth}
        \centering
        \includegraphics[width=\textwidth]{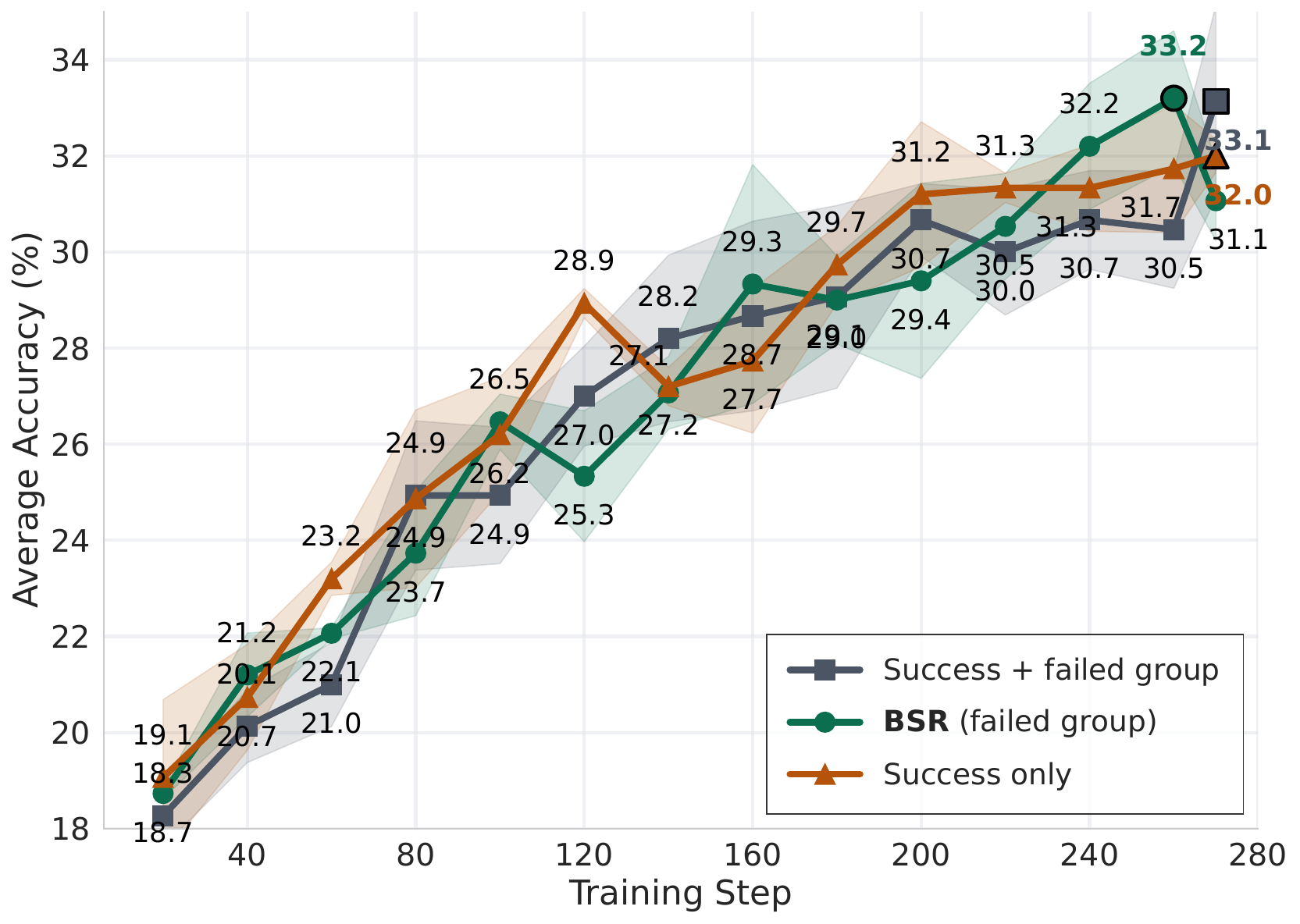}
        \caption{Average accuracy.}
        \label{fig:group-ablation-avg}
    \end{subfigure}
    \hfill
    \begin{subfigure}{0.48\textwidth}
        \centering
        \includegraphics[width=\textwidth]{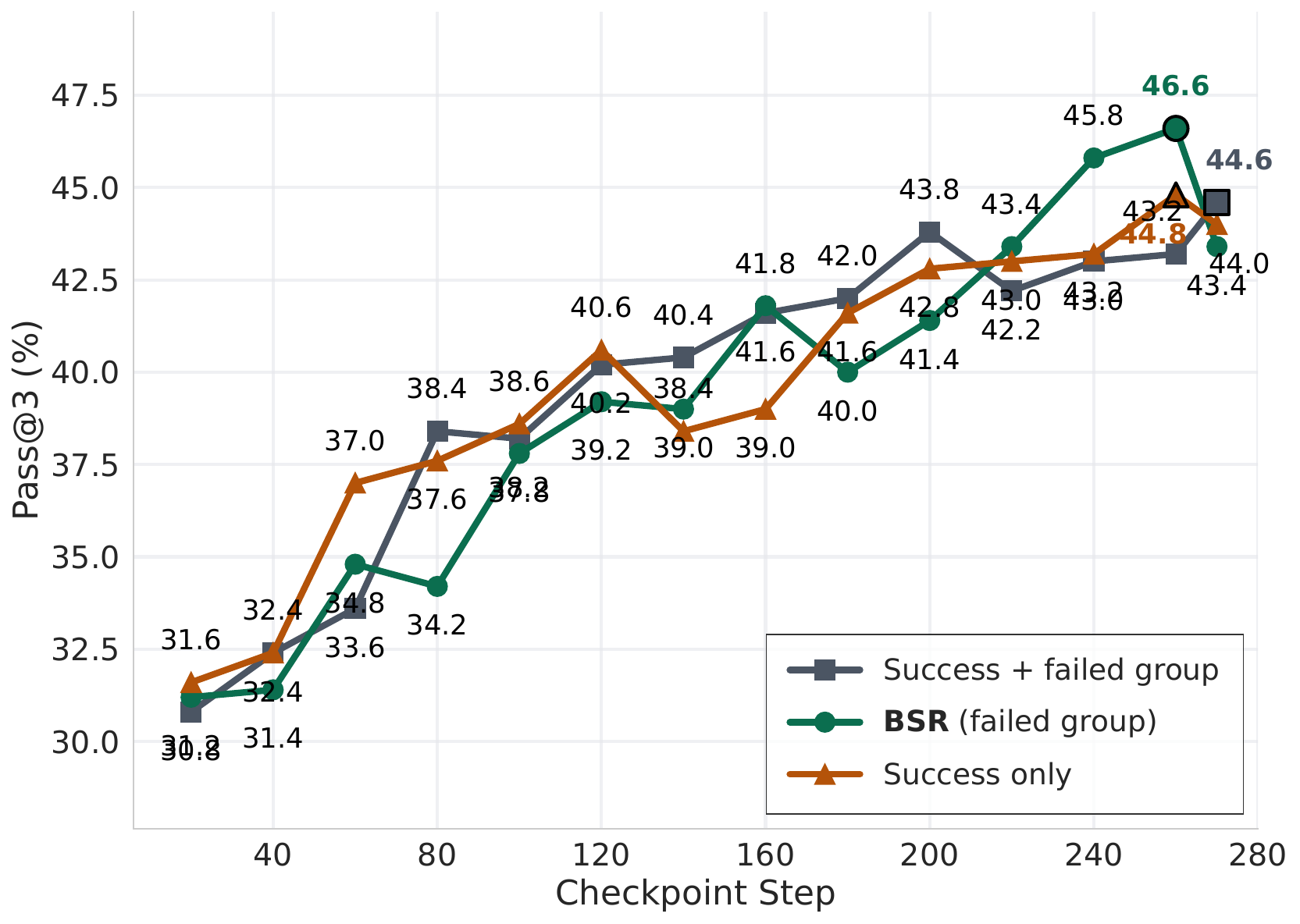}
        \caption{Pass@3.}
        \label{fig:group-ablation-pass3}
    \end{subfigure}

    \caption{Verifier-outcome-bucket ablation on SWE-bench Verified while training Qwen3-14B. We compare applying BSR to failed trajectories only, successful trajectories only, and both successful and failed trajectories.}
    \label{fig:group-ablation}
\end{figure}

\begin{figure}[htbp]
    \centering

    \begin{subfigure}{0.48\textwidth}
        \centering
        \includegraphics[width=\textwidth]{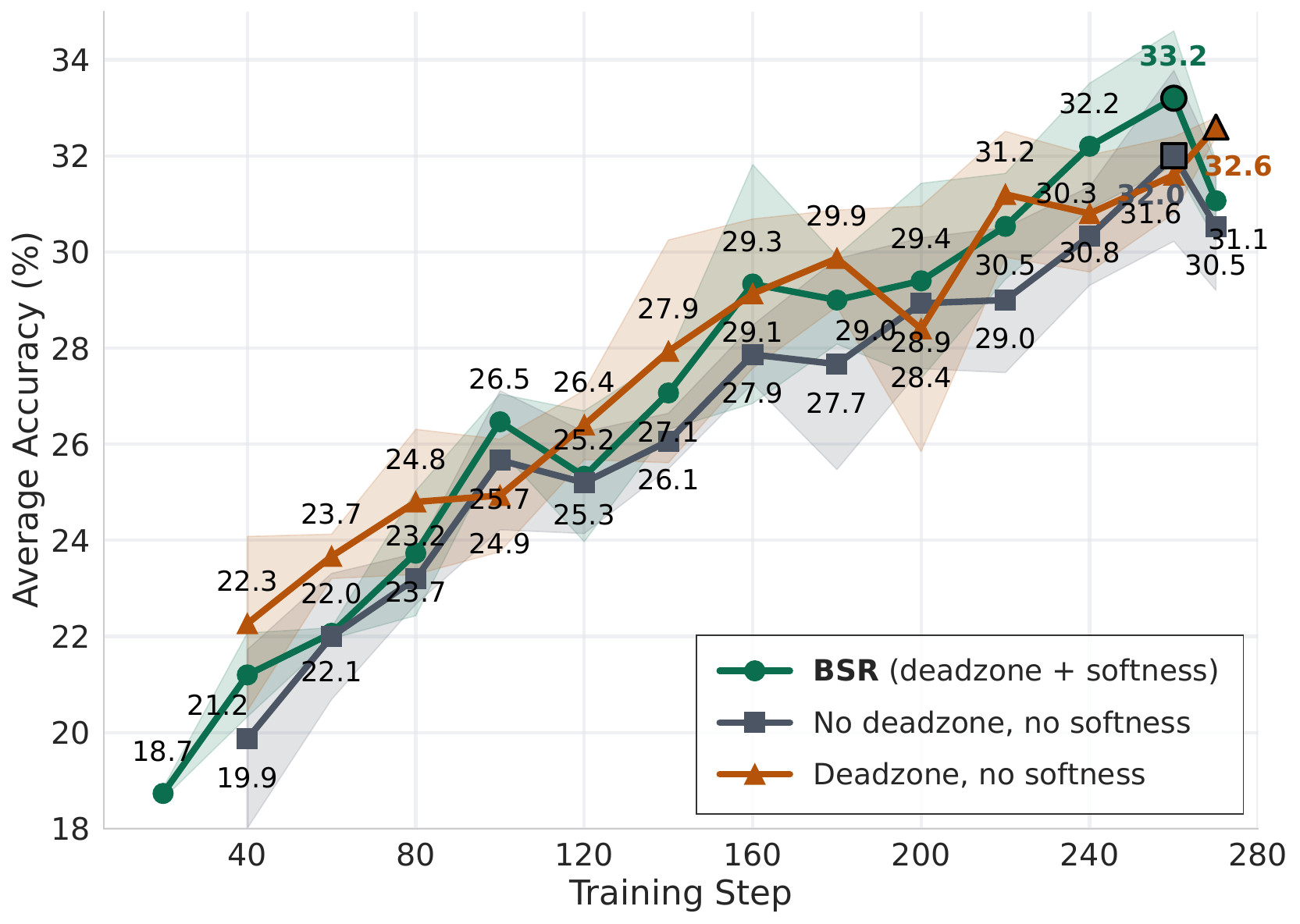}
        \caption{Average accuracy.}
        \label{fig:deadzone-ablation-avg}
    \end{subfigure}
    \hfill
    \begin{subfigure}{0.48\textwidth}
        \centering
        \includegraphics[width=\textwidth]{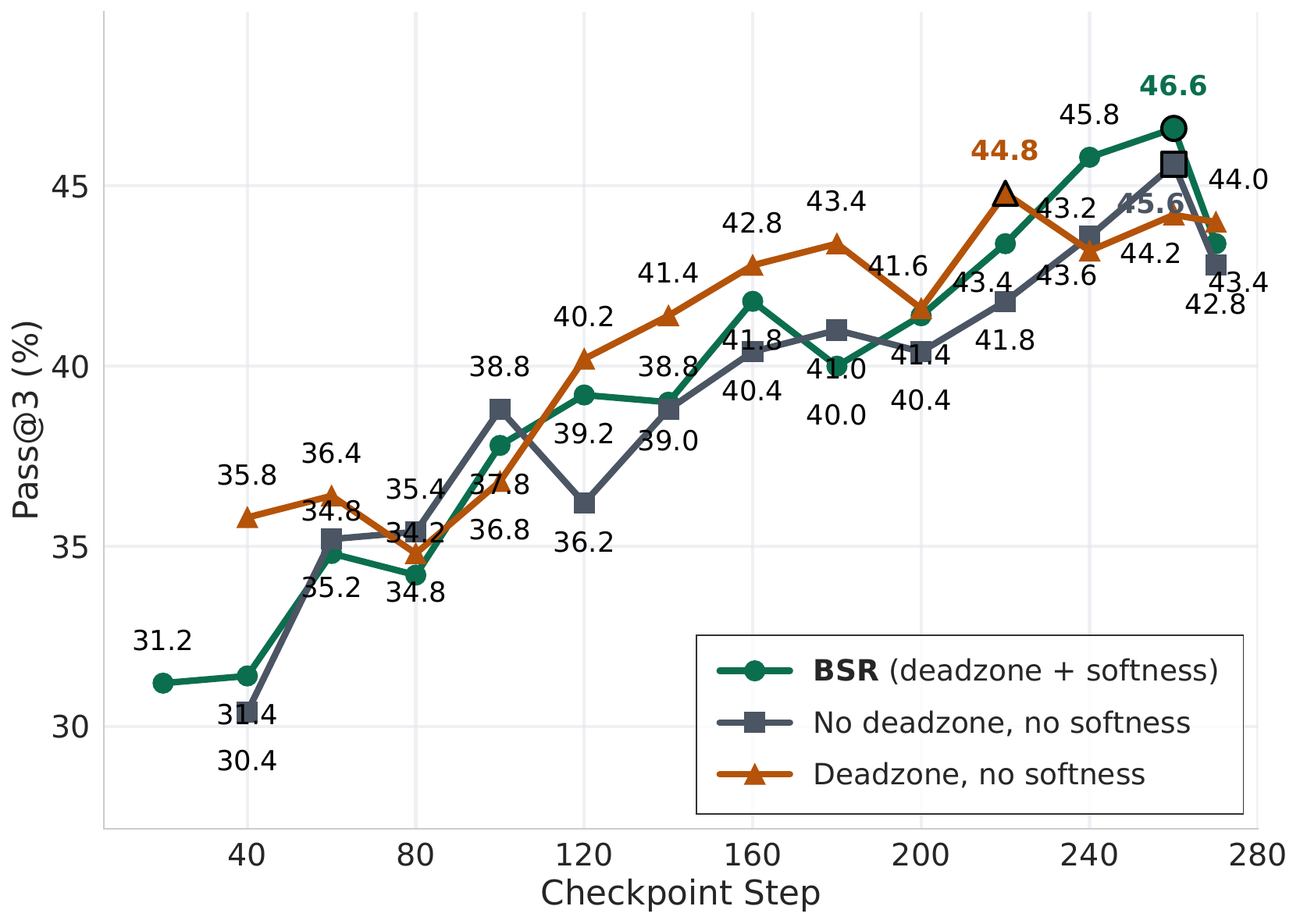}
        \caption{Pass@3.}
        \label{fig:deadzone-ablation-pass3}
    \end{subfigure}

    \caption{Reward-mapping ablation on SWE-Bench Verified while training Qwen3-14B. We compare the full BSR reward mapping against variants without the deadzone and/or softness.}
    \label{fig:deadzone-ablation}
\end{figure}

% \begin{figure}[t]
%     \centering
%     \begin{subfigure}{0.48\textwidth}
%         \centering
%         \includegraphics[width=\linewidth]{figs/pairwise_eval_accuracy_overall.pdf}
%         \caption{Overall accuracy.}
%         \label{fig:judgment-alignment-accuracy}
%     \end{subfigure}
%     \hfill
%     \begin{subfigure}{0.48\textwidth}
%         \centering
%         \includegraphics[width=\linewidth]{figs/pairwise_eval_exact_accuracy_overall.pdf}
%         \caption{Exact accuracy.}
%         \label{fig:judgment-alignment-exact-accuracy}
%     \end{subfigure}
% \caption{Additional judgment-agreement metrics over training.
% We compare the Qwen3-14B model with Kimi K2.5 on the same sampled
% pairwise comparisons used in Figure~\ref{fig:judgment-alignment}.}
%     \label{fig:judgment-alignment-appendix}
% \end{figure}

\subsection{Additional Evaluation on Terminal-Bench 2.1}
\label{app:terminalbench}

We additionally evaluate BSR on Terminal-Bench 2.1. Each configuration is evaluated on 89 tasks with three independent evaluation trials. Table~\ref{tab:terminalbench-results} reports average accuracy across the three trials and Pass@3, the fraction of tasks solved by at least one trial. BSR consistently improves both average accuracy and Pass@3 across all evaluated model sizes and training budgets, demonstrating that the gains extend beyond the primary evaluation setting.

\subsection{Additional Results}

We provide additional training curves in this section. Figure~\ref{fig:critic-score} shows the mean scores during training for both Qwen3-14B and Qwen3-32B.
Figure~\ref{fig:14b-checkpoint-comparison} presents the full 400-step curves on SWE-Bench Verified for Qwen3-14B, while Figure~\ref{fig:32b_labeled_all} shows the detailed training curves for Qwen3-32B.
Figures~\ref{fig:group-ablation} and~\ref{fig:deadzone-ablation} provide the full results on SWE-Bench Verified for the verifier-outcome-bucket ablation and reward-mapping ablations, respectively. 

% Finally, Figure~\ref{fig:judgment-alignment-appendix} reports additional judgment-alignment metrics over training, complementing the macro-F1 analysis in Figure~\ref{fig:judgment-alignment}.

\begin{table*}[t]
\centering
\small
\begin{tabular}{lcc}
\toprule
Statistic & Baseline & BSR (Ours) \\
\midrule
\multicolumn{3}{l}{\textbf{Process metrics}} \\
Reference file viewed before first edit $\uparrow$ & 0.733 & \textbf{0.748} \\
Search-to-edit conversion $\uparrow$ & 0.465 & \textbf{0.494} \\
Post-edit test rate $\uparrow$ & 0.301 & \textbf{0.311} \\
Duplicate command rate $\downarrow$ & 0.209 & \textbf{0.208} \\
\midrule
\multicolumn{3}{l}{\textbf{Basic workload statistics}} \\
Average total steps & 28.90 & 28.42 \\
Average first edit step & 5.33 & 5.24 \\
Average number of searches & 4.74 & 4.06 \\
Average number of reads & 5.39 & 5.56 \\
Average number of edits & 12.46 & 12.16 \\
Average number of tests & 0.747 & 0.797 \\
Average number of post-edit tests & 0.746 & 0.797 \\
Average number of errors & 13.38 & 12.76 \\
Average unique files read & 1.97 & 2.11 \\
Average unique files edited & 3.02 & 3.17 \\
Average search result files per search & 7.20 & 7.12 \\
Average observation lines per step & 32.21 & 31.85 \\
Average total time (s) & 1132.28 & 1068.55 \\
Average total tokens & 18283.74 & 18572.82 \\
\midrule
Trajectory count & 1500 & 1500 \\
\bottomrule
\end{tabular}
\caption{Appendix trajectory behavior statistics over the full set of trajectories. Arrows indicate the preferred direction for process metrics. Basic workload statistics are descriptive and should be interpreted in context.}
\label{tab:appendix-trajectory-behavior}
\end{table*}

\subsection{Trajectory Behavior Statistics}
\label{app:trajectory_behavior_stats}

We analyze 1,500 evaluation trajectories per method on SWE-bench
Verified from Qwen3-14B training logs and report both workload statistics and debugging-process
metrics.

\paragraph{Metrics Used.}
We report two groups of trajectory-level statistics. The first group contains basic workload statistics, including the number of steps, first edit step, numbers of search/read/edit/test actions, post-edit tests, errors, unique files read and edited, runtime, and token usage. These statistics describe how the agent spends computation, but are not always directly comparable: for example, fewer steps may indicate efficiency or premature stopping, while more searches may indicate either careful investigation or unfocused exploration.

The second group contains process metrics that more directly reflect debugging behavior. \emph{Reference file viewed before first edit} measures whether the agent localized a reference-relevant file before editing. \emph{Search-to-edit conversion} measures whether search results led to downstream edits. \emph{Post-edit test rate} measures whether the agent verified its changes after editing. \emph{Duplicate command rate} captures repeated commands. We treat these process metrics as the primary behavioral indicators, while workload statistics provide supporting context.

\paragraph{Behavioral Differences.}
Table~\ref{tab:appendix-trajectory-behavior} shows that BSR improves several process metrics associated with investigation and verification. Compared with the baseline, BSR more often views a reference-relevant file before the first edit (0.748 vs.\ 0.733), has higher search-to-edit conversion (0.494 vs.\ 0.465), and performs more post-edit testing (0.311 vs.\ 0.301). It also slightly reduces duplicate commands (0.208 vs.\ 0.209). The workload statistics are consistent with this shift: BSR uses fewer steps, performs fewer searches, reads slightly more, runs more tests, encounters fewer errors, and completes trajectories faster.

\paragraph{Summary.}
BSR shows stronger reference-oriented investigation, search-to-edit
follow-through, and post-edit verification. These aggregate patterns
are consistent with changes in investigation and verification behavior,
with workload statistics providing complementary evidence about how the
agent allocates its debugging effort.

\begin{figure}[h]
    \centering
    \includegraphics[width=0.7\linewidth]{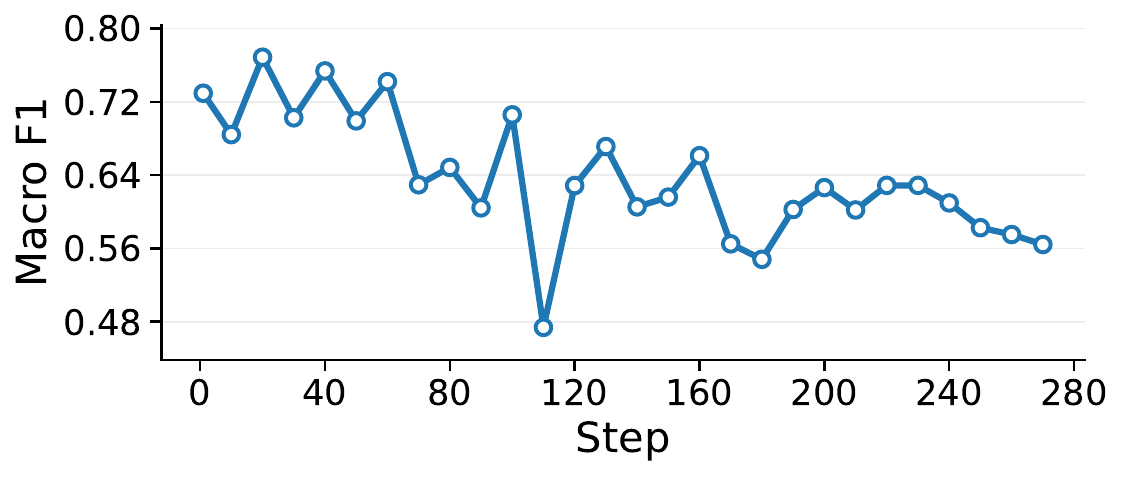}
    \caption{Agreement with Kimi-2.5 on sampled pairwise trajectory
    comparisons over training, measured by macro F1. At each evaluation
    point, 200 on-policy comparison pairs are sampled and re-evaluated
    using Kimi-2.5 as an external reference judge.}
    \label{fig:judgment-alignment}
\end{figure}

\subsection{Judge Quality Investigation}

We evaluate the policy's pairwise judgments by measuring agreement with
Kimi K2.5~\citep{team2026kimi} on 200 sampled comparisons every 10
training steps. As shown in Figure~\ref{fig:judgment-alignment}, the
self-judge maintains meaningful agreement with the external judge
throughout training, with macro F1 generally around 0.6--0.7. Although
agreement decreases somewhat at later checkpoints, it remains
substantial and the corresponding BSR models continue to improve in
downstream performance. This suggests that the policy's judgments
remain useful for reward refinement, while improving the stability of
self-judgment over training is an interesting direction for future work.

% \section{Infrastructure Setup and Adjustments}

% \paragraph{How to setup kub}
% \begin{itemize}
%     \item Kub setup and caching: pitfall: set the max pod and threshold to delete
%     \item Bounded Requests for maximum pod run
% \end{itemize}
% \paragraph{Environment Pre-fetching}
% \paragraph{Training Time}

\begin{figure*}
    \centering
    \small
    \begin{llmprompt}

You are a strict evaluator for software engineering agents.
You will be shown two candidate trajectories produced by two agents solving the same programming task. Each trajectory contains the agent’s actions, and outcomes.\\

Each trajectory may also end with a compact stats block. Attend to those metrics as supporting evidence about efficiency and behavior, but do not treat them as the sole basis for judgment. 

Both trajectories ultimately succeeded and produced a correct patch that passes all tests.

Your goal: compare the two trajectories and decide which one demonstrates better debugging and code-editing behavior that, if reinforced during training, is more likely to improve future task accuracy. 

Use a rubric-based pairwise judgment. Evaluate only observable behavior in the trajectories. Do not judge writing quality.

For correct-vs-correct comparisons, prefer the trajectory whose behavior is more efficient, causally grounded, and safer to reinforce for future generalization.\\

Rubric for correct trajectories:\\
1. Failure-surface investigation:\\
- Did it investigate the right failure surface?\\
- Did it look in the right code paths, files, or behaviors?\\
2. Evidence-based diagnosis:\\
- Did it use evidence to narrow uncertainty before changing code?\\
- Did it run useful checks or inspections to distinguish between plausible causes?\\
3. Targeted and efficient execution:\\
- Were actions and edits targeted and efficient?\\
- Did it avoid unnecessary exploration, wasted motion, or overly broad changes?
4. Root-cause alignment:\\
- Did the patch address the root cause rather than only symptom-matching?
- Was the reasoning connected to why the fix should work?\\
5. Safety and regression awareness:\\
- Did it preserve existing behavior and avoid likely regressions?\\
- Did it avoid risky or unnecessary edits?\\

Useful trajectory stats to attend to when they are present:\\
- num\_bash\_runs\_before\_last\_edit and num\_bash\_runs\_after\_last\_edit: useful for checking whether the agent validated before changing code and whether it verified after making edits.\\
- num\_repeated\_actions: repeated identical actions are usually a negative signal.\\
- num\_edits\_without\_prior\_view: edits made without first viewing the file are usually a negative signal.\\

Ignore:\\
- explanation style\\
- verbosity\\
- confidence\\
- stylistic fluency\\
- superficial similarity to a known patch unless it directly reflects better debugging behavior\\

Always pick one winner: "A" or "B".\\
Then assign a coarse strength-of-preference score:\\
- 1 = slight\\
- 2 = clear\\
- 3 = strong\\

Strength should reflect how much better the winner is in this pair, not absolute trajectory quality.\\

Output format (first give your thinking and then the result):\\
<think>your analysis here</think>\\
Winner: A or B \\
Strength: 1 or 2 or 3 
    \end{llmprompt}
    \caption{System Prompt for Success-Bucket Pairwise Comparison.}
    \label{prompt:success}
\end{figure*}

\begin{figure*}
    \centering
    \small
    \begin{llmprompt}

You are a strict evaluator for software engineering agents.
You will be shown two candidate trajectories produced by two agents solving the same programming task. Each trajectory contains the agent’s actions, and outcomes.\\

Each trajectory may also end with a compact stats block. Attend to those metrics as supporting evidence about efficiency and behavior, but do not treat them as the sole basis for judgment.

Both trajectories ultimately failed and produced a wrong patch that fails one or more tests.

Your goal: compare the two trajectories and decide which one demonstrates better debugging and code-editing behavior that, if reinforced during training, is more likely to improve future task accuracy.

Use a rubric-based pairwise judgment. Evaluate only observable behavior in the trajectories. Do not judge writing quality. 

For wrong-vs-wrong comparisons, prefer the trajectory that demonstrates more useful bug-finding behavior and is closer to eventual success, even though it did not solve the task.\\

Rubric for wrong trajectories:\\
1. Progress toward the real bug:\\
- Did it make progress toward the real bug, even if it failed?\\
- Did it move closer to the relevant code path, mechanism, or failure cause?\\
2. Evidence gathering and discriminative checks:\\
- Did it gather useful evidence or run discriminative checks?\\
- Did it reduce uncertainty in a meaningful way?\\
3. Directional quality of edits and actions:\\
- Were edits directionally sensible rather than random or cargo-culted?\\
- Were commands, inspections, and modifications plausibly useful?\\
4. Relevance of search path:\\\
- Did it avoid digging deeper into irrelevant paths?\\
- Did it avoid getting stuck in unproductive loops or distractions?\\
5. Informative and recoverable failure:\\
- Did it fail in an informative and recoverable way?\\
- Did it avoid introducing extra damage, confusion, or regressions?\\

Useful trajectory stats to attend to when they are present:\\
- num\_bash\_runs\_before\_last\_edit and num\_bash\_runs\_after\_last\_edit: useful for checking whether the agent gathered evidence before editing and tested or probed after edits.\\
- num\_repeated\_actions: repeated identical actions are usually a negative signal.\\
- num\_edits\_without\_prior\_view: edits made without first viewing the file are usually a negative signal.\\

Ignore:\\
- explanation style\\
- verbosity\\
- confidence\\
- stylistic fluency\\
- superficial similarity to a known patch unless it directly reflects better debugging behavior\\

Always pick one winner: "A" or "B".\\
Then assign a coarse strength-of-preference score:\\
- 1 = slight\\
- 2 = clear\\
- 3 = strong\\

Strength should reflect how much better the winner is in this pair, not absolute trajectory quality.\\

Output format (first give your thinking and then the result):\\
<think>your analysis here</think>\\
Winner: A or B \\
Strength: 1 or 2 or 3 \\
    \end{llmprompt}
    \caption{System Prompt for Failure-Bucket Pairwise Comparison.}
    \label{prompt:fail}
\end{figure*}

\begin{figure*}
    \centering
    \small
    \begin{llmprompt}
Below are two candidate agents solving the same programming task. 

GitHub Issue to be addressed:\\
\{task\}\\
\{golden\_patch\_section\}\\

\-\-\-\-\-\-\-\-\-\-
Trajectory A:\\
\{traj\_a\}\\

\-\-\-\-\-\-\-\-\-\-

Trajectory B:\\
\{traj\_b\}\\
    \end{llmprompt}
    \caption{User Prompt for ranking.}
    \label{prompt:user}
\end{figure*}

\begin{figure*}
    \centering
    \tiny
    \begin{llmprompt2}
<think>
Okay, let's compare the two trajectories. Both agents tried to solve the same problem but ended up with wrong patches. The goal is to determine which one showed better debugging and code-editing behavior.\\

First, looking at Trajectory A. The agent started by exploring the file structure, found the relevant multi.py file, and searched for 'get\_indexer' to locate the method. They then viewed specific code sections to understand where the method was used. After several steps, they added a new 'get\_indexer' method in the MultiIndex class to check for level mismatches. They created a test script and ran it, which raised the correct error. The stats show that they did one bash run after the last edit, which is good because it tests the change. There was one edit without prior view, but overall, their steps were methodical, focusing on the right files and methods.\\

Now, Trajectory B. The agent also explored the files, found the same multi.py and base.py. They tried modifying the base.py's get\_indexer method to check for MultiIndex levels. However, they hit an import error because they changed the import statements to relative imports, which caused a circular dependency. They tried multiple times to fix the import issues, leading to repeated actions and errors. The stats show many repeated actions (11) and some edits without prior viewing. The agent's approach led to more errors and didn't resolve the import problem, even after several attempts.\\

Comparing the two, Trajectory A made progress toward the real bug by correctly identifying the MultiIndex's get\_indexer method and adding the check there. Their edits were directional and sensible, leading to a test that showed the error. Trajectory B, while attempting to modify the base class, introduced import issues and struggled with circular dependencies, leading to more errors and less progress toward the actual solution. Even though both failed, A's approach was closer to the correct fix and had more effective debugging steps.\\
</think>\\

**Reasoning:**  \\
Trajectory A demonstrated more effective debugging and code-editing behavior despite both failing.  \\

1. **Progress toward the real bug:**  \\
   - Trajectory A directly identified the `MultiIndex` class in `multi.py` and focused on the `get\_indexer` method, which is the correct location for the fix. They added a method to check for level mismatches, which aligns with the issue description. \\ 
   - Trajectory B attempted to modify the `get\_indexer` method in the base `Index` class but introduced a circular import error due to incorrect relative imports, which deviated from the actual problem (which requires a `MultiIndex`-specific fix). 
2. **Evidence gathering and discriminative checks:**  \\
   - Trajectory A systematically searched for `get\_indexer` in `multi.py` and examined code context to locate the relevant method. They created a test script and ran it to validate their change, which produced the expected error.  \\
   - Trajectory B made multiple edits to `base.py` without fully understanding the import structure, leading to repeated errors (e.g., `ImportError`). Their approach lacked discriminative checks to isolate the root cause.  \\
3. **Directional quality of edits and actions:**  \\
   - Trajectory A’s edits were focused on the `MultiIndex` class and its `get\_indexer` method, which is the correct code path. Their final edit added a clear, targeted check for level mismatches.  \\
   - Trajectory B’s edits were less directional. They modified the `Index` class’s `get\_indexer` method, which is not the correct location for `MultiIndex`-specific logic. Their repeated attempts to fix import issues showed poor progress. \\ 
4. **Relevance of search path:**  \\
   - Trajectory A avoided irrelevant paths (e.g., `base.py` was only explored for context, not as the primary fix location).  \\
   - Trajectory B repeatedly edited `base.py` and `multi.py` without resolving the circular import, getting stuck in a loop of unproductive fixes.  \\
5. **Informative and recoverable failure:**  \\
   - Trajectory A’s failure was recoverable: their test script raised the expected `TypeError`, confirming their logic was correct. The failure was due to incomplete implementation (e.g., not handling all edge cases).  \\
   - Trajectory B’s failure was due to a cascading import error that they could not resolve, which was less informative and harder to recover from.  \\
**Conclusion:**  \\
Trajectory A’s approach was more methodical, targeted, and closer to the correct solution. Their edits were meaningful and validated through testing, while Trajectory B’s attempts were hindered by circular dependencies and less relevant modifications.

Winner: A\\
Strength: 3  \\

    \end{llmprompt2}
    \caption{Example Ranking Response from Qwen3-14B.}
    \label{example_compare}
\end{figure*}